\pdfoutput=1

\documentclass[11pt]{article}

\usepackage[preprint]{acl}

\usepackage{times}
\usepackage{latexsym}

\usepackage[T1]{fontenc}

\usepackage[utf8]{inputenc}

\usepackage{microtype}

\usepackage{inconsolata}

\usepackage{graphicx}

\usepackage{placeins} 
\usepackage{enumitem}
\usepackage{minted}
\usepackage{booktabs}
\usepackage[table]{xcolor}
\definecolor{okabe_red}{HTML}{D55E00}
\definecolor{okabe_blue}{HTML}{0072B2}
\definecolor{okabe_green}{HTML}{009E73}
\definecolor{okabe_orange}{HTML}{E69F00}
\definecolor{okabe_purple}{HTML}{CC79A7}
\definecolor{tab_blue}{HTML}{3182BD}
\definecolor{tab_blue2}{HTML}{6BAED6}
\definecolor{tab_blue3}{HTML}{9ECAE1}
\definecolor{tab_orange}{HTML}{E6550D}
\definecolor{tab_orange2}{HTML}{FD8D3C}
\definecolor{tab_orange3}{HTML}{FDAE6B}
\definecolor{tab_green}{HTML}{31A354}
\definecolor{tab_green2}{HTML}{74C476}
\definecolor{tab_green3}{HTML}{A1D99B}
\definecolor{tab_purple}{HTML}{756BB1}
\definecolor{tab_gray}{HTML}{636363}
\definecolor{tab_lightblue}{HTML}{EAF2F8}
\definecolor{tab_lightorange}{HTML}{FCEEE6}
\definecolor{tab_lightgreen}{HTML}{EAF6EE}
\definecolor{tab_lightgray}{HTML}{E0E0E0}

\usepackage{tabularray}
\usepackage{amsmath}
\usepackage{stmaryrd,graphicx} 
\usepackage{subcaption}
\usepackage{amssymb}
\usepackage{pifont}
\usepackage{listings}
\usepackage{xfrac}

\newcommand{\footlabel}[2]{%
    \addtocounter{footnote}{1}%
    \footnotetext[\thefootnote]{%
        \addtocounter{footnote}{-1}%
        \refstepcounter{footnote}\label{#1}%
        #2%
    }%
    $^{\ref{#1}}$%
}

\newcommand{\myfootref}[1]{%
    $^{\ref{#1}}$%
}

\newcounter{normalfootc}
\renewcommand{\footnote}[1]{%
    \footlabel{footsaferefiwontuse\thenormalfootc}{#1}%
    \addtocounter{normalfootc}{1}%
}

\newcommand{\cmark}{\ding{52}}%
\newcommand{\xmark}{\textcolor{lightgray}{\ding{55}}}%

\makeatletter
\newcommand{\fixed@sra}{$\vrule height 2\fontdimen22\textfont2 width 0pt\shortrightarrow$}
\newcommand{\shortarrow}[1]{%
  \mathrel{\text{\rotatebox[origin=c]{\numexpr#1*45}{\fixed@sra}}}
}
\makeatother

\newlength{\fancyvrbtopsep}
\newlength{\fancyvrbpartopsep}
\makeatletter
\FV@AddToHook{\FV@ListParameterHook}{\topsep=\fancyvrbtopsep\partopsep=\fancyvrbpartopsep}
\makeatother
\title{Survey Response Generation:\\Generating Closed-Ended Survey Responses In-Silico\\with Large Language Models}

\author{
  \textbf{Georg Ahnert\textsuperscript{1}},
  \textbf{Anna-Carolina Haensch\textsuperscript{2,3,4}},
  \textbf{Barbara Plank\textsuperscript{2,3}},
  \textbf{Markus Strohmaier\textsuperscript{1,5,6}}
\\
  \textsuperscript{1}University of Mannheim;
  \textsuperscript{2}LMU Munich;
  \textsuperscript{3}Munich Center for Machine Learning;
\\
  \textsuperscript{4}University of Maryland, College Park;
  \textsuperscript{5}GESIS Cologne;
  \textsuperscript{6}CSH Vienna
\\
  \small{
    \textbf{Correspondence:} \href{mailto:georg.ahnert@uni-mannnheim.de}{georg.ahnert@uni-mannnheim.de}
  }
}

\begin{document}
\maketitle
\begin{abstract}
Many \textit{in-silico} simulations of human survey responses with large language models (LLMs) focus on generating closed-ended survey responses, whereas LLMs are typically trained to generate open-ended text instead.
Previous research has used a diverse range of methods for generating closed-ended survey responses with LLMs, and a standard practice remains to be identified.
In this paper, we systematically investigate the impact that various \textbf{Survey Response Generation Methods} have on predicted survey responses. We present the results of 32 mio.\@ simulated survey responses across 8 Survey Response Generation Methods, 4 political attitude surveys, and 10 open-weight language models.
We find \textbf{significant differences} between the Survey Response Generation Methods in both individual-level and subpopulation-level alignment.
Our results show that Restricted Generation Methods perform best overall, and that reasoning output does not consistently improve alignment.
Our work underlines the significant impact that Survey Response Generation Methods have on simulated survey responses, and we develop practical recommendations on the application of Survey Response Generation Methods.

\end{abstract}

\begin{figure}[ht!]
    \centering
    \includegraphics[width=0.9\linewidth]{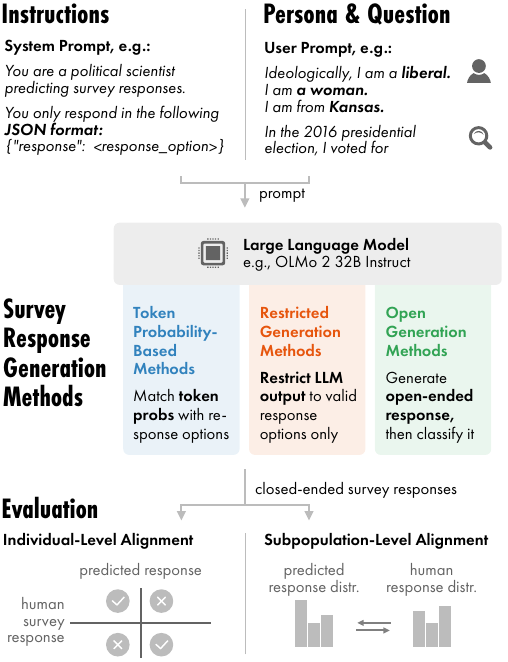}
    \caption{\textbf{Survey Response Generation Methods Elicit Closed-Ended Survey Responses From LLMs.} We prompt all models with a combined Persona \& Question Prompt to predict political attitudes in the U.S. or Germany. All implemented Survey Response Generation Methods elicit closed-ended survey responses from the LLMs we investigate. We evaluate the individual-level alignment of these responses against human survey data, and the distribution alignment in subpopulations against human response distributions.}
    \label{fig:figure1}
\end{figure}

\section{Introduction}
A growing body of research simulates human survey responses by prompting large language models (LLMs) to answer survey questions \citep[][inter alia]{argyle_out_2023}.
While generative LLMs are designed to generate open-ended text, previous studies have implemented various approaches to constraining LLMs to closed-ended survey responses~\cite{ma_potential_2024}.
We define \textbf{Survey Response Generation Methods} as techniques used to elicit closed-ended responses from large language models to survey questions on attitudes, opinions, and values.
Previous research has shown that the closed-ended responses of an LLM can vary strongly from its open-ended responses~\citep{rottger_political_2024, wang_my_2024}, but a standard Survey Response Generation Method for social simulations with LLMs has not yet been identified.

\begin{table*}[t!]
    \centering
    \small
    \begin{tblr}{
      colsep=4pt,
      colspec={Q[l, font=\bfseries] Q[1.5mm, colsep=0pt] Q[l, font=\bfseries] X[c] X[c] X[c] X[c] X[c] }, 
      rowsep=1pt,
      width=\textwidth,
      hline{2}={black},
      hborder{3-9}={belowspace=1.5pt},
      hborder{2,5,8}={belowspace=2.5pt},
      hline{3-9}={3-8}{tab_lightgray, 0.01pt},
      hline{5,8}={black, 0.3pt},
      vline{4-8}={black},
    }
\SetCell[c=3]{} & & & {Accesses\\Token-\\Probabilities} & {Enforces\\Format w/\\Instructions} & {Restricts\\LLM\\Vocabulary} & {Generates\\Open Out-\\put First\textsuperscript{1}} & {Generates\\Probability\\Distribution}\\
\SetCell[r=3]{bg=tab_lightblue, fg=tab_blue} {Token Prob.-\\Based Methods} & \SetCell{bg=tab_blue} & First-Token Probabilities  & \cmark & \cmark & \xmark & \xmark & \cmark \\
& \SetCell{bg=tab_blue2} & First-Token Restricted    & \cmark & \cmark & \cmark & \xmark & \cmark \\
& \SetCell{bg=tab_blue3} & Answer Prefix             & \cmark & \cmark & \cmark & \xmark & \cmark \\
\SetCell[r=3]{bg=tab_lightorange, fg=tab_orange} {Restricted\\Generation\\Methods} & \SetCell{bg=tab_orange} & Restricted Choice        & \xmark & \cmark & \cmark & \xmark & \xmark \\
& \SetCell{bg=tab_orange2} & Restricted Reasoning    & \xmark & \cmark & \cmark & \cmark & \xmark \\
& \SetCell{bg=tab_orange3} & Verbalized Distribution & \xmark & \cmark & \cmark & \xmark & \cmark \\
\SetCell[r=2]{bg=tab_lightgreen, fg=tab_green} {Open Generation\\Methods} & \SetCell{bg=tab_green} & Open-Ended Classification & \xmark & \xmark\textsuperscript{2} & \xmark\textsuperscript{2} & \cmark & \xmark \\
& \SetCell{bg=tab_green2} & Open-Ended Distribution  & \xmark & \xmark\textsuperscript{2} & \xmark\textsuperscript{2} & \cmark & \cmark \\
    \end{tblr}
    \caption{\textbf{Overview of Survey Response Generation Methods.} Based on previous research~\cite[][inter alia]{ma_potential_2024, rottger_political_2024}, we implement a range of Token Probability-Based Methods~{\color{tab_blue}$\blacksquare$}, Restricted Generation Methods~{\color{tab_orange}$\blacksquare$}, and Open Generation Methods~{\color{tab_green}$\blacksquare$} for the production of closed-ended survey responses with LLMs. \textsuperscript{1}With reasoning models, all methods generate open-ended reasoning output first. \textsuperscript{2}The LLM is unrestricted at first, but restricted in its vocabulary and through formatting instructions in the second, classification step.}
    \label{tab:methods_overview}
\end{table*}

In this paper, we evaluate 8 diverse Survey Response Generation Methods against 4 human survey datasets on both individual-level and subpopulation-level alignment, as shown in Figure~\ref{fig:figure1}.
We focus our evaluation on the \textbf{simulation of human survey responses on political attitudes in the US and Germany} and replicate the findings of three influential studies~\citep{argyle_out_2023, von_der_heyde_vox_2025, santurkar_whose_2023} while including novel Survey Response Generation Methods.
We find that Restricted Generation Methods perform best, and that reasoning output does not improve performance. Instructing the model to verbalize probabilities for all response options consistently yields the best distributional alignment.

With this study, we contribute to the literature in three ways: (i) we present results from \textbf{extensive evaluations} of Survey Response Generation Methods with diverse survey datasets, prompt perturbations, LLMs, and decoding parameters for a total of 32 mio.\@ simulated survey responses. (ii) We highlight the \textbf{significant impact of Survey Response Generation Methods} on simulated survey responses, and (iii) we develop \textbf{practical recommendations} on which Survey Response Generation Method to use.

\section{Survey Response Generation Methods}

A growing body of research uses LLMs to simulate human survey responses to questions on attitudes, opinions, and values \textit{in-silico}~\cite[][inter alia]{argyle_out_2023, park_generative_2024, boelaert_machine_2025}. In these studies, an LLM is provided with a description of a persona and a survey question and prompted to predict the survey response of this individual. A large fraction of human survey data is available in a closed-ended format, i.e., each question has a set of response options, categorical or ordinal. Researchers, thus, deploy a diverse range of Survey Response Generation Methods for generating closed-ended survey responses with LLMs.

\subsection{Token Probability-Based Methods} \label{subsec:token_methods}
This set of Survey Response Generation Methods assumes access to token probabilities and that there are tokens that uniquely identify a single response option. For instance, the token \textit{``Don''} would encode \textit{Donald Trump} as a response option in a question on 2016 U.S. vote choice. We extract its token probability directly from the model output. To increase robustness, the probabilities of tokens that encode the same response option are added up---for instance, \textit{``donald'', ``Tru'',} etc.

The \textbf{First-Token Probabilities Method}~{\color{tab_blue}$\blacksquare$} is a popular implementation of this approach that extracts probabilities directly on the first output token that a model generates~\cite[][inter alia]{argyle_out_2023, dominguez-olmedo_questioning_2024, santurkar_whose_2023, holtdirk_learning_2025}.

Since many LLM inference providers only return the top $k$ output tokens and not probabilities over the whole vocabulary, it could happen that some response options do not get a token probability assigned.\footnote{e.g., the \href{https://developers.openai.com/api/reference/resources/chat/subresources/completions/methods/create}{OpenAI API} returns the top 20 tokens only.} We therefore additionally implement the \textbf{First-Token Restricted Method}~{\color{tab_blue2}$\blacksquare$} that restricts the vocabulary of an LLM to only output tokens from the possible response options.

\citet{wang_my_2024} showed that the first-token probability of an LLM might not always align with its open-ended response, which instead may start with a prefix, e.g.,\textit{``My answer is ''}. One potential mitigation would be to consider token probabilities only after a fixed response-prefix. We implement the prefix above in the \textbf{Answer Prefix Method}~{\color{tab_blue3}$\blacksquare$}.

\subsection{Restricted Generation Methods} \label{subsec:restricted_methods}

This set of methods uses formatting instructions in the system prompt of an LLM to obtain an output that can be easily parsed~\citep[][inter alia]{hartmann_political_2023, motoki_more_2023}. We additionally restrict the vocabulary of the LLM to only the valid response options. While the latter is not strictly necessary for these methods, it ensures that the model follows the formatting instructions provided.

For the \textbf{Restricted Choice Method}~{\color{tab_orange}$\blacksquare$}, we dynamically define a JSON schema for each survey question that forces a simple JSON format and only allows for valid response options to be generated by the LLM inside this JSON as follows:

\vspace{5pt}
{
\footnotesize
\noindent
\texttt{\phantom{+}\{"answer\_option": <a valid response option>\}}
}
\vspace{5pt}

The \textbf{Restricted Reasoning Method}~{\color{tab_orange2}$\blacksquare$} extends the previous method by first forcing the model to generate reasoning before it generates its choice of response option. The resulting JSON is formatted as follows:

\vspace{5pt}
{
\footnotesize
\noindent
\texttt{\phantom{+}\{"reasoning": <any string>,\\\phantom{++}"answer\_option": <a valid response option>\}}
}
\vspace{5pt}

While the previous two methods force the model to generate a single response for each prediction, the \textbf{Verbalized Distribution Method}~{\color{tab_orange3}$\blacksquare$} restricts the model to generate a probability distribution over all response options, following~\citet{meister_benchmarking_2025}. The resulting JSON is formatted as follows:

\vspace{5pt}
{
\footnotesize
\noindent
\texttt{\phantom{+}\{<response\_option\_A>: <probability>,\\\phantom{++}<response\_option\_B>: <probability>, ... \}}
}
\vspace{5pt}

Note that we still obtain a distribution over all response options per individual.

\subsection{Open Generation Methods} \label{subsec:open_methods}
These Survey Response Generation Methods are inspired by a line of work which argues that LLM evaluations with survey questions should be based on open-ended LLM responses \citep{rottger_political_2024, wright_llm_2024, myrzakhan_open-llm-leaderboard_2024}. Our goal is different, as we aim to simulate human survey responses instead of evaluating the LLM itself~\citep[see also][]{sorensen_position_2024}. Still, for the Open Generation Methods, we do not restrict the output of the model in any way. Instead, we obtain an open-ended response from each LLM and, in a second step, prompt the same model to classify this output according to the survey question and response options at hand. 
The \textbf{Open-Ended Classification Method}~{\color{tab_green}$\blacksquare$} implements this two-step approach using the Restricted Choice Method for the classification step, i.e., classifying the output for a single selected response option. The \textbf{Open-Ended Distribution Method}~{\color{tab_green2}$\blacksquare$} uses the Verbalized Distribution Method for the classification step, respectively.

\section{Experimental Setup}

\begin{table}[b!]
    \centering
    \scriptsize
    \begin{tblr}{
      width=\linewidth,
      colspec={X | Q Q Q Q | Q Q},
      colsep=4pt,
      rowsep=1pt,
      column{3,6}={rightsep=1pt},
      column{4,7}={leftsep=1pt},
      cell{2-5}{2} = {l}, 
      row{1}={font=\bfseries},
      column{1}={font=\bfseries},
      hline{2}={-}{},
      hborder{2}={belowspace=3pt},
    }
    Survey & {\#Indivi-\\duals\textsuperscript{1}} & \SetCell[c=2]{}{\#Questions \&\\Topic} & & {Lang. \&\\Country}  & \SetCell[c=2]{}{\#Options \&\\Scale Type} & \\
    ANES 2016 & 4270 & 1: & vote choice                    & EN, US & \nobreak\hspace{.43em}3: & categorical \\
    GLES 2017 & 1976 & 1: & vote choice                    & DE, DE & \nobreak\hspace{.43em}9: & categorical \\
    GLES 2025 & 6771 & 1: & vote choice                    & DE, DE & 10: & categorical \\ 
    ATP 2021  & 500\textsuperscript{2} & 7: & {social \& cul-\\tural change} & EN, US & \nobreak\hspace{.43em}5: & {ordinal\\(Likert-scale)}
    \end{tblr}
    \caption{\textbf{Evaluation Datasets.} We simulate political attitudes across multiple languages, countries, and response scales. \textsuperscript{1}Excluding individuals with missing data on the simulated survey questions. \textsuperscript{2}Random sample out of 10221 participants in wave 92.}
    \label{tab:datasets}
\end{table}

\subsection{Datasets}
In this paper, we present a comprehensive evaluation of the above described Survey Response Generation Methods and \textbf{compare the obtained predictions to survey responses from human participants.}
We focus on political attitudes, as this has been a popular subject for in-silico surveys in the past. Our evaluation spans multiple countries, two languages, and several response option scales, as shown in Table~\ref{tab:datasets}.
We predict vote choice from the 2016 American National Election Study~\citep{anes_2016_2016}, replicating study 2 from \citet{argyle_out_2023} with an English-language prompt. We further replicate \citet{von_der_heyde_vox_2025}'s study on vote choice in Germany, predicting responses from the 2017 German Longitudinal Study~\citep{gles_gles_2017} with a German language prompt. As these two datasets might have leaked into the training data of state-of-the-art LLMs, we also predict self-reported vote choice from the 2025 German federal election~\citep{gles_gles_2025}.\footnote{All Llama 3 models and the OLMo training data was published before the survey fieldwork period, and the Qwen models shortly thereafter, so it is unlikely that survey results from the 2025 German election have leaked into the models.} Finally, we partially\footlabel{ATP}{Partial replication of selected questions and with only 1 random seed as we implement multiple, more computationally expensive Survey Response Production Methods.} replicate \citet{santurkar_whose_2023} by simulating questions 48--54 from wave 92 of the American Trends Panel~\citep{atp_american_2021}.

\subsection{Prompt Design}
To perform predictions for the ANES 2016 and GLES 2017/2025 datasets, we include the same persona attributes and use the same prompt format as \citet{argyle_out_2023}, and \citet{von_der_heyde_vox_2025}, only adjusting the year for GLES 2025.
For the ATP 2021 dataset, we use \citet{santurkar_whose_2023}'s ``bio'' format, which is closest to the other prompts.
We run all evaluations with the following system prompt: \textit{``You are a political scientist predicting responses to the following question:…''}, and include formatting instructions for the Token Probability-Based and Restricted Generation Methods. All prompt templates are provided in Appendix Section~\ref{sec:instruction_prompts}.

\subsection{Response Option Scales}
To investigate the robustness of all Survey Response Generation Methods against prompt rephrasing, we include 4 variants of the response option scale. The \textbf{Full Text} variants consist of the full text of each response option, without using an index---e.g., \textit{`Clinton', `Trump', `Non-voter'} for a question on 2016 U.S. vote choice. The \textbf{Indexed} variants, sometimes called \textit{multiple choice questions} \citep{balepur_which_2025}, use an index for the response options instead---e.g., \textit{`A', `B', `C'}. Additionally, we evaluate the prompt both in the original order of response options and in a \textbf{reversed} order, as previous work has shown option order to be a major source of output variation \citep{tjuatja_llms_2024, rupprecht_prompt_2025}.

\subsection{Language Models}
We perform our evaluations on \textbf{10 open-weight, instruction-tuned and reasoning models} of different sizes from the Llama 3~\citep[3B, 8B, 70B;][]{llama_team_llama_2024}, OLMo 2~\citep[1B, 7B, 32B;][]{olmo_team_2_2025}, and Qwen 3~\cite[8B, 32B;][]{qwen_team_qwen3_2025} families of models---see Appendix Table~\ref{app_tab:language_models} for the specific model IDs. We also include Qwen 3 8B and Qwen 3 32B with enabled reasoning output in our evaluations. We compare responses obtained from  greedy decoding and from the model default temperatures for 3 random seeds.
\myfootref{ATP} For the Open Generation Methods, we scale temperature for the first, open-ended response step, but keep the default temperature of the model for the classification step.
We evaluate the first-token probabilities of reasoning models at the first token after the reasoning output.
For computational details, see Appendix~\ref{sec:computational_details}.

\subsection{Evaluation}
We evaluate all Survey Response Generation Methods by comparing the generated survey responses to human survey data.
We include a \textbf{stratified baseline} obtained from randomly shuffling the human survey responses in each dataset. As upper bounds of \textbf{achievable alignment} on the individual level, we obtain predictions from cross-validations of tuned random forest models. For achievable alignment on the subpopulation level, we use repeated \sfrac{1}{3} sub-sampling~\citep[following][]{suh_language_2025}. We do not expect scores beyond these bounds since human survey responses are not fully predictable.

\paragraph{Individual-Level Alignment}
First, we calculate the macro avg.\ F1-score of the generated survey responses against the individual human survey responses.
For all methods that generate an individual-level distribution across all response options (see Table~\ref{tab:methods_overview}), we select the most probable response option for evaluation.

\begin{figure*}[t!]
    \centering
    \includegraphics[width=\linewidth]{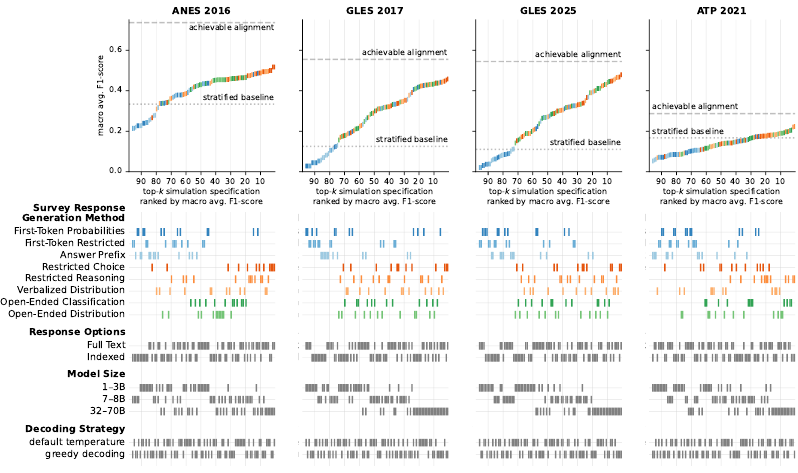}
    \caption{\textbf{Individual-Level Alignment Between In-Silico Generated and Human Survey Responses by Dataset (Columns) and Simulation Specification.}
    \textbf{Top:} macro avg.\ F1-score $(\uparrow)$ for each aggregated simulation specification, mean across the respective runs. \textbf{Bottom:} simulation specification---Survey Response Generation Method, response option variant, model size, and decoding strategy---sorted by macro avg.\ F1-score $(\rightarrow)$.
    Invalid responses are counted as incorrect. \textbf{Individual-level alignment varies strongly between Survey Response Generation Methods.}
    For subpopulation-level alignment, see Appendix Figures~\ref{app_fig:all_datasets_tv_specification} \& \ref{app_fig:all_datasets_dcor_specification}.
    }
    \label{fig:all_models_f1_ANES}
\end{figure*}

\paragraph{Subpopulation-Level Alignment}
Second, we evaluate the alignment of responses on a subpopulation-level by aggregating individual responses. This is different from previous research that simulated subpopulations directly~\citep[e.g.,][]{santurkar_whose_2023}. However, simulating individual survey responses first is more versatile as it enables, e.g., individual-level imputation of missing human survey data.

We split the set of respondents into subpopulations by considering all unique values of all persona attributes that were originally included in the simulation by \citet{argyle_out_2023}, \citet{von_der_heyde_vox_2025}, and \citet{santurkar_whose_2023}, e.g., women \& men, people from different states, etc. For \textit{age}, we construct age brackets by flooring to multiples of 10.
For all methods that do not generate an individual-level distribution across all response options (see Table~\ref{tab:methods_overview}), we create said distributions through one-hot encoding. We normalize all individual-level distributions to sum up to 1.
We then calculate the distribution over response options for a subpopulation as the mean across individual-level distributions.
We report the subpopulation-level alignment between the generated distributions and the distribution found in the human survey data for the respective subpopulation using total variation distance for categorical response options (ANES/GLES) and 1-Wasserstein distance for ordinal response options (ATP 2021).



\section{Individual-Level Alignment}

First, we view survey response generation as a prediction task and evaluate individual-level alignment. Figure~\ref{fig:all_models_f1_ANES} shows macro average F1-scores (top) of aggregated simulation specifications (bottom)---
method, response options variant, model size, and decoding strategy.

\textbf{Survey Response Generation Methods have a large impact on individual-level alignment.} Even for specifications that surpass the stratified baseline, we see a difference of $>0.35$ on the GLES 2017 and GLES 2025 datasets. Token-Probability Based Methods can yield comparable F1-scores for larger models, but perform poorly for smaller models. This is partially because these methods are more prone to generate invalid responses, especially for reasoning models (see Appendix Figure~\ref{app_fig:invalid_responses}). Restricted Generation Methods, and in particular the Restricted Choice Method, perform best across most datasets and approach the upper bound of achievable alignment on some. LLMs with more parameters also generally outperform smaller models, but we observe no clear pattern for response option scales and decoding strategies.

To investigate the specific impact of Survey Response Generation Methods on individual-level alignment, we fit OLS regressions on each dataset. Table~\ref{tab:f1_ols} shows the regression coefficients of all Survey Response Generation Methods compared to the First-Token Probabilities Method as a reference, as it is one of the most popular methods. We find that \textbf{the Restricted Choice Method leads to significant improvements in individual-level alignment,} followed by the Restricted Reasoning Method and the Open-Ended Classification Method. The Verbalized Distribution Method and the Open-Ended Distribution Method yield significant improvements for some datasets, even if they are designed to generate probability distributions across all response options rather than a single response. We observe similar patterns when using \textit{accuracy} as a metric (see Appendix Table~\ref{tab:accuracy_ols}). Open Generation Methods generally improve individual-level alignment, but show smaller coefficients for most datasets. This indicates that long, open-ended ``reasoning'' does not improve the results in our task, while also being orders of magnitude less computationally efficient (see Appendix Figure~\ref{app_fig:GPU_time}).

\begin{table}[t!]
    \centering
    \footnotesize
    \begin{tblr}{
      colsep=3pt,
      colspec={Q[1.5mm, colsep=0pt] Q[l, font=\bfseries, colsep=4pt] | X[l] X[l] X[l] X[l]}, 
      rowsep=2pt,
      row{1,2}={font=\bfseries, rowsep=0.5pt},
      width=\linewidth,
      hline{3,4,6,9}={black},
      hborder{3,4,6,9}={belowspace=3pt},
    }
& & ANES & GLES & GLES & ATP \\
& & 2016 & 2017 & 2025 & 2021 \\
\SetCell{bg=tab_blue} & Intercept & .343* & .037* & .050 & .107* \\
\SetCell{bg=tab_blue2} & First-Token Restricted & .082* & .051* & .066 & -.032* \\
\SetCell{bg=tab_blue3} & Answer Prefix & .013 & .069* & .059 & -.021* \\
\SetCell{bg=tab_orange} & Restricted Choice & \textbf{.148*} & \textbf{.242*} & \textbf{.218*} & .015 \\
\SetCell{bg=tab_orange2} & Restricted Reasoning & \underline{.138*} & .219* & .196* & \textbf{.052*} \\
\SetCell{bg=tab_orange3} & Verbalized Distribution & .118* & \underline{.233*} & \underline{.216*} & .011 \\
\SetCell{bg=tab_green} & Open-Ended Classif. & .127* & .215* & .184* & \underline{.037*} \\
\SetCell{bg=tab_green2} & Open-Ended Distrib. & .114* & .212* & .177* & .018* \\
    \end{tblr}
    \caption{\textbf{Regression Coefficients for Individual-Level Alignment $(\uparrow)$.}
    OLS regression for each dataset with macro avg.\ F1-score $(\uparrow)$ in each simulation specification as the dependent variable. We use Survey Response Generation Method, response option scale, and LLM as independent variables. We show coefficients for the Survey Response Generation Methods (Reference: First-Token Probabilities~{\color{tab_blue}$\blacksquare$}) and include additional coefficients in Appendix Table~\ref{app_tab:f1_ols_full}. Highest coefficient in \textbf{bold}, second highest \underline{underlined.} \textbf{The Restricted Choice Method~{\color{tab_orange}$\blacksquare$} leads to a significant improvements.} $\text{*}\,p < 0.05$, Benjamini-Hochberg adjusted.}
    \label{tab:f1_ols}
\end{table}

\subsection{Individual-Level Robustness}

In addition to accuracy, we evaluate the robustness of the generated survey responses against prompt perturbations on an individual level. We show the mean agreement between the response option scales---Full Text / Indexed, original / reversed---in Table~\ref{tab:robustness_allDatasets}.
We observe that especially smaller models (1--3B parameters) often generate disagreeing survey responses. Across all model sizes, Token Probability-Based Methods show little agreement. This indicates that they are subject to biases against certain response options (e.g., A-bias). Out of the Survey Response Generation Methods we evaluate, Restricted Choice and Open-Ended Classification yield the highest agreement across all model sizes, and the Restricted Reasoning Method as well as Open-Ended Distribution are generally not far off.

On one hand, individual-level robustness could be desirable, as the generated survey responses should not be influenced by perturbations of the prompt or of the response option scales. On the other hand, perfect agreement could not be desirable, as persona prompts only provide limited information about the individuals they aim to simulate, and human survey responses are often not uniquely predictable given these attributes. This is exemplified by difficult-to-predict cases.


\begin{table}[t!]
    \centering
    \footnotesize
    \begin{tblr}{
      colsep=3pt,
      colspec={Q[1.5mm, colsep=0pt] Q[l, font=\bfseries, colsep=4pt] | X[c] X[c] X[c]}, 
      rowsep=2pt,
      row{1,2}={font=\bfseries, rowsep=0.5pt},
      width=\linewidth,
      hline{3,6,9}={black},
      hborder{3,4,6,9}={belowspace=3pt},
      vline{4,5}={2-Z}{black}
    }
& & \SetCell[c=3]{c} Model Size & & \\
& & 1--3B & 7--8B & 32--70B \\
\SetCell{bg=tab_blue} & First-Token Probabilities & 0.34 & 0.23 & 0.27 \\
\SetCell{bg=tab_blue2} & First-Token Restricted & 0.07 & 0.15 & \underline{0.53} \\
\SetCell{bg=tab_blue3} & Answer Prefix & 0.11 & 0.24 & 0.32 \\
\SetCell{bg=tab_orange} & Restricted Choice & 0.18 & \underline{0.49} & \textbf{0.74} \\
\SetCell{bg=tab_orange2} & Restricted Reasoning & 0.15 & \underline{0.48} & \textbf{0.69} \\
\SetCell{bg=tab_orange3} & Verbalized Distribution & 0.07 & \underline{0.41} & \textbf{0.60} \\
\SetCell{bg=tab_green} & Open-Ended Classif. & 0.20 & \underline{0.54} & \textbf{0.76} \\
\SetCell{bg=tab_green2} & Open-Ended Distrib. & 0.14 & \underline{0.45} & \textbf{0.69} \\
    \end{tblr}
    \caption{\textbf{Individual-Level Robustness Across Scales.} Mean Fleiss's $\kappa\,(\uparrow)$ across all datasets, results with more than 10\% invalid values are excluded. We \underline{underline} $\kappa \geq0.4$ and use \textbf{bold font} for $\kappa \geq 0.6$. \textbf{Token Probability-Based Methods {\color{tab_blue}$\blacksquare$}{\color{tab_blue2}$\blacksquare$}{\color{tab_blue3}$\blacksquare$} show poor individual-level robustness across response option scales.} Separate results for each dataset and model, as well as agreement across random seeds are shown in Appendix Figure~\ref{app_fig:robustness_seeds_scales}.}
    \label{tab:robustness_allDatasets}
\end{table}

\begin{table}[b!]
    \centering
    \scriptsize
    \begin{tblr}{
        width={\linewidth},
        colspec={Q Q Q[c] Q[c] | X[colsep=5pt] X },
        colsep=3.2pt,
        rowsep=1pt,
        row{1}={font=\bfseries},
    }
{political\\ideology} & {party\\identification}  & {US\\state} & ... & {true\\vote choice} & {predicted\\vote choice} \\ 
\hline \hborder{belowspace=3pt}
               & a strong Democrat   & CA & ... & Trump     & Clinton   \\ 
extr. conserv. & a strong Republican & TX & ... & Clinton   & Trump     \\ 
conservative   & Indep. leaning Rep. & AZ & ... & Clinton   & Trump     \\ 
liberal        & Indep. leaning Dem. & OH & ... & Trump     & Clinton   \\ 
conservative   & a strong Republican & NJ & ... & Non-Voter & Trump     \\ 
    \end{tblr}
    \caption{\textbf{Most Difficult to Predict Cases in the ANES 2016 dataset,} as identified by a calibrated logistic regression with out-of-fold predictions obtained from 5-fold cross-validation. All five predictions have a true class probability of $\approx0$. See Appendix~\ref{sec:instruction_prompts} for a full list of persona attributes included in the simulation. Given the limited information available, we would not expect an LLM to correctly predict the vote choice reported by these individuals in the ANES survey.}
    \label{tab:hard_cases}
\end{table}

\section{Subpopulation-Level Alignment}
While manually investigating the individuals who are the most difficult to predict, we found that their reported vote choice often runs counter to what we would intuitively expect given their other attributes (see Table~\ref{tab:hard_cases}). Evaluating Survey Response Generation Methods with individual-level alignment only does not account for such cases, so we also consider response distributions in subpopulations.

We again fit an OLS regression model using TLV and 1-Wasserstein metrics as the dependent variables. We show the coefficients for the Survey Response Generation Methods in Table~\ref{tab:tv_ols}.
We see that \textbf{subpopulation-level alignment varies significantly across the Survey Response Generation Methods,} even if simulation specifications that yield many invalid responses are excluded.
We find that the \textbf{Verbalized Distribution Method generates well-aligned survey responses} across all datasets.
The Restricted Reasoning Method and the Open-Ended Classification Method also yield good subpopulation-level alignment, even if they generate a single survey response and not a distribution over response options for each individual.

Token Probability-Based Methods, on the other hand, often perform worse.
This could be explained by Restricted Generation Methods being closer to benchmark-like tasks that LLMs have been trained on during instruction-tuning, while token probabilities are not well aligned for instruction-tuned LLMs---see also \citet{hu_simbench_2025} for a similar argument around this \textit{alignment-simulation tradeoff}.
Regression coefficients for \textit{Jensen-Shannon divergence} as a dependent variable are presented in Appendix Table~\ref{tab:jsd_ols} and show similar trends.



\subsection{Reasoning Models} \label{subsec:reasoning_results}

For both individual-level and subpopulation-level, we find that methods that do not perform open-ended ``reasoning''---especially the Verbalized Distribution Method---can yield well-aligned survey responses.
This prompts further investigation of two dedicated reasoning models: Qwen 3 8B \& Qwen 3 32B. We observe that \textbf{reasoning output does not consistently improve subpopulation-level alignment,} and in some cases even degrades alignment---e.g., for the best performing Survey Response Generation Method on the GLES 2025 dataset for Qwen 3 8B: Restricted Reasoning (see Appendix Figure~\ref{fig:reasoning_models_align}).
This is in line with previous findings which show that chain-of-thought reasoning mostly improves model output on mathematical and logic tasks~\citep{sprague_cot_2024}.
Our evaluation is, however, limited to the default reasoning traces that Qwen 3 8B \& Qwen 3 32B are trained to produce and future research could investigate more structured or theory-informed reasoning strategies.

\begin{table}[t!]
    \centering
    \footnotesize
    \begin{tblr}{
      colsep=3pt,
      colspec={Q[1.5mm, colsep=0pt] Q[l, font=\bfseries, colsep=4pt] | X[l] X[l] X[l] | X[l]}, 
      row{1}={font=\bfseries},
      rowsep=2pt,
      width=\linewidth,
      hline{2,3,5,8}={black},
      hborder{2,3,5,8}={belowspace=3pt}
    }
& & {ANES\\2016} & {GLES\\2017} & {GLES\\2025} & {ATP\\2021} \\
\SetCell{bg=tab_blue} & Intercept & .322* & .476* & .578* & .070* \\
\SetCell{bg=tab_blue2} & First-Token Restricted & .105* & .044 & -.069 & .008 \\
\SetCell{bg=tab_blue3} & Answer Prefix & .049 & -.120* & -.202* & .002 \\
\SetCell{bg=tab_orange} & Restricted Choice & .045 & -.165* & -.287* & .012* \\
\SetCell{bg=tab_orange2} & Restricted Reasoning & .061* & -.197* & \underline{-.312*} & \underline{-.010*} \\
\SetCell{bg=tab_orange3} & Verbalized Distribution & \textbf{-.028} & \textbf{-.219*} & -.296* & \textbf{-.016*} \\
\SetCell{bg=tab_green} & Open-Ended Classif. & .072* & -.174* & -.306* & -.001 \\
\SetCell{bg=tab_green2} & Open-Ended Distrib. & .038 & \underline{-.216*} & \textbf{-.319*} & -.006 \\
    \end{tblr}
    \caption{\textbf{Regression Coefficients for Subpopulation-Level Alignment $(\downarrow)$.}
    OLS regression for each dataset, with total variation distance $(\downarrow)$ as the dependent variable for the ANES and GLES datasets and 1-Wasserstein distance $(\downarrow)$ for the ATP 2021 dataset. Results with more than 10\% invalid values were excluded.  We use Survey Response Generation Method, response option variant, and LLM as independent variables. We show coefficients for Survey Response Generation Methods (Reference: First-Token Probabilities~{\color{tab_blue}$\blacksquare$}) and include additional coefficients in Appendix Table~\ref{app_tab:tv_ols_full}. \textbf{Verbalized Distribution~{\color{tab_orange3}$\blacksquare$} leads to significant improvements.} $\text{*}\,p < 0.05$, Benjamini-Hochberg adjusted.}
    \label{tab:tv_ols}
\end{table}

\subsection{A Global Perspective On Subpopulation-Level Alignment}

Total variation distance and 1-Wasserstein distance compare \textit{in-silico} generated survey responses with human survey responses separately in each subpopulation. The results from all subgroups then have to be aggregated by using a (weighted) average or a regression model, as shown in Table~\ref{tab:tv_ols}. \textit{Distance correlation} is a measure of dependence between random vectors that enables an alternative, global perspective on subpopulation-level alignment~\citep{szekely_measuring_2007} and has previously been used to compare human to model judgment distributions~\citep{chen_seeing_2024}. High distance correlation indicates that the dependency structure between subpopulations and survey response distributions in the generated data resembles that in human survey data. We calculate the distance correlation for each simulation specification and show the results in the rightmost column of Table~\ref{tab:aggregate_ols} alongside aggregated results for previously discussed individual-level and subpopulation-level assessments.
Again, we find that the Restricted Generation Methods yield good global alignment.

\begin{table}[t!]
    \centering
    \footnotesize
    \begin{tblr}{
      colsep=2pt,
      colspec={Q[1.5mm, colsep=0pt] Q[l, font=\bfseries, colsep=2pt] | X[c] X[c] | X[c] X[c]}, 
      row{1,2}={font=\bfseries},
      rowsep=2pt,
      width=\linewidth,
      hline{3,4,6,9}={black},
      hborder{3,4,6,9}={belowspace=3pt},
      vline{4,6}={2-Z}{black},
    }
& & \SetCell[c=2]{}{Individual-\\Level} & & \SetCell[c=2]{}{Subpop.-\\Level} & \\
& & {Align-\\ment} & {Robu-\\stness} & {Align-\\ment} & {Global\\Align.} \\
\SetCell{bg=tab_blue} & Intercept & -1.647* & -1.183* & -0.930* & -1.150* \\
\SetCell{bg=tab_blue2} & First-Token Restrict. & 0.751* & 0.461* & -0.029 & 0.613* \\
\SetCell{bg=tab_blue3} & Answer Prefix & 0.415 & 0.009 & 0.667* & -0.073 \\
\SetCell{bg=tab_orange} & Restricted Choice & \underline{1.535*} & \underline{1.178*} & 0.845* & 1.138* \\
\SetCell{bg=tab_orange2} & Restricted Reasoning & \textbf{1.576*} & 1.097* & 1.090* & \textbf{1.302*} \\
\SetCell{bg=tab_orange3} & Verbalized Distrib. & 1.443* & 0.655* & \textbf{1.433*} & 1.104* \\
\SetCell{bg=tab_green} & Open-Ended Classif. & 1.440* & \textbf{1.382*} & 0.882* & \underline{1.281*} \\
\SetCell{bg=tab_green2} & Open-Ended Distrib. & 1.273* & 0.930* & \underline{1.097*} & 1.104* \\
    \end{tblr}
    \caption{\textbf{Regression Coefficients Across All Surveys (normalized, $\uparrow$).} OLS regressions for individual-level alignment (macro avg.\ F1-score), robustness (Fleiss' $\kappa$), subpopulation-level alignment ($1 -$ total variation distance / 1-Wasserstein distance) and global alignment (distance correlation). Results from each metric are z-score normalized separately per dataset and results with more than 10\% invalid values were excluded. We use dataset, Survey Response Generation Method, response option variant, and LLM as independent variables. We show coefficients for Survey Response Generation Methods (Reference: First-Token Probabilities~{\color{tab_blue}$\blacksquare$}) and include additional coefficients in Appendix Table~\ref{app_tab:aggregate_ols_full}. \textbf{Restricted Generation Methods~{\color{tab_orange}$\blacksquare$}{\color{tab_orange2}$\blacksquare$}{\color{tab_orange3}$\blacksquare$} consistently yield significant improvements} and are computationally efficient. $\text{*}\,p < 0.05$, Benjamini-Hochberg adjusted.}
    \label{tab:aggregate_ols}
\end{table}

\section{Related Work}

\paragraph{Generating Survey Responses with LLMs}


Previous research has investigated how perturbations of the prompt impact the responses of a model to survey questions \cite[][inter alia]{tjuatja_llms_2024, dominguez-olmedo_questioning_2024, rupprecht_prompt_2025}.
For instance, \citet{mcilroy-young_order-independence_2024} investigated option-order effects, i.e., the tendency of LLMs to respond to survey questions differently when the order in which the response options are presented is changed.
Recently, \citet{cummins_threat_2025} demonstrated that the survey responses generated by LLMs can vary widely across input-text specifications and models.
In this paper, we instead focus on closed-ended output of LLMs, and still find significant differences between simulation specifications. Future research should combine both prompts and Survey Response Generation Methods for a holistic assessment of how LLMs generate survey responses.

Few papers have investigated the specific impact of Survey Response Generation Methods so far.
\citet{wang_my_2024} identified that the most probable first output token often does not match the open-ended responses of an LLM when prompted with survey questions. Through our assessments, we go beyond dissimilarities in the generated responses and identify the Survey Response Generation Method that works best for a given survey and LLM.
\citet{meister_benchmarking_2025} compared different methods for generating response distributions and found that the Verbalized Distribution Method works best, which is why we also implement it. Our work goes further, as we simulate the responses of individual survey participants instead of subpopulations directly. We evaluate Survey Response Generation Methods on non-English datasets (GLES 2017, GLES 2025), and compare a wider range of methods, including Open Generation Methods.

\paragraph{Response Generation in Other Settings}
Another line of research has investigated closed-ended and open-ended model responses in other contexts.
\citet{rottger_political_2024} found that adding instructions on the response format, and forcing models to, e.g., `take a clear stance', alters the response option that a model chooses. \citet{tam_let_2024} also observed a negative impact of format instructions for mathematical reasoning tasks. Finally, \citet{licht_measuring_2025} evaluate methods for annotating scalar constructs with LLMs and see improvements with pairwise comparisons and token probability-weighted scores. We extend this line of research to the simulation of survey responses, including persona prompts and human survey data for comparison.


\section{Recommendations and Conclusion}

In this paper, we present a \textbf{systematic assessment of Survey Response Generation Methods} for generating closed-ended survey responses \textit{in-silico} with large language models.
Our evaluations span 8 Survey Response Generation Methods, 4 political attitude datasets across 2 countries and languages, and 10 open-weight LLMs, as well as multiple robustness checks.\footnote{Code \& Data: \url{https://github.com/dess-mannheim/survey_response_generation}}

Recommendations: (i) We argue that the \textbf{choice of Survey Response Generation Method should be well-justified} for \textit{in-silico} surveys since we find significant differences between these methods. (ii) We \textbf{do not recommend the use of Token Probability-Based Methods~{\color{tab_blue}$\blacksquare$}{\color{tab_blue2}$\blacksquare$}{\color{tab_blue3}$\blacksquare$},} as they generate misaligned survey responses. (iii) For predicting closed-ended survey responses, we suggest to \textbf{consider Restricted Generation Methods~{\color{tab_orange}$\blacksquare$}{\color{tab_orange2}$\blacksquare$}{\color{tab_orange3}$\blacksquare$} first,} as they consistently showed significant improvement over other methods while also being more computationally efficient than Open Generation Methods~{\color{tab_green}$\blacksquare$}{\color{tab_green2}$\blacksquare$}.

This paper should serve as a starting point for future research on how to generate valid, reliable, and useful survey responses with LLMs.

\newpage
\section*{Limitations}

Our main focus in this paper lies on the evaluation of Survey Response Generation Methods for the \textit{in-silico} simulation of surveys on political attitudes. We include a diverse range of datasets across countries with different political systems and languages and aim to replicate influential studies on \textit{in-silico} surveys~\citet{argyle_out_2023, von_der_heyde_vox_2025, santurkar_whose_2023}. Many of our overall findings generalize across these datasets. Still, further evaluation should be performed in non-Western contexts and on surveys of attitudes, opinions, and values that go beyond the topics that we have studied. The already large variety of simulation specifications we have included---Survey Response Generation Methods, LLM, response option scale, decoding strategy, etc.---also created computational constraints that did not allow us to investigate, for instance, the impact of language and country/political context independently between the ANES and GLES datasets.

Future research should also investigate further perturbations to the response options scales such as a missing midpoint option, which have been found to impact human and LLM survey responses~\cite{tjuatja_llms_2024, rupprecht_prompt_2025, mcilroy-young_order-independence_2024}. Other applications of closed-ended questions, e.g., for LLM benchmarking or to investigate model alignment, are beyond the scope of this paper, and it remains to be demonstrated whether our results generalize to these contexts as well.

We define Survey Response Generation Methods as being concerned with how closed-ended survey responses are generated by and can be extracted from LLMs. This specifically limits the scope of our paper to exclude different approaches to prompting LLMs for survey simulation~\citep{lutz_prompt_2025, park_generative_2024}. Instead, we adopt a persona prompt template from each study that we replicate~\citep{argyle_out_2023, von_der_heyde_vox_2025, santurkar_whose_2023}. For each model and dataset, we evaluate all Survey Response Generation Methods using 4 response option scale variants, $\{\text{Full Text, Indexed}\}\times\{\text{original order, reversed order}\}$, and 3 random seeds\footnote{For the ATP 2021 dataset, we only used 1 seed but include responses to 7 different questions.} to investigate the robustness of our findings. Still, alternative persona prompts (e.g., interview-style) have been shown to positively influence the LLM response alignment \citep{lutz_prompt_2025} and might interact differently with different Survey Response Generation Methods.
Finally, while we do investigate the joint impact of \texttt{temperature} and \texttt{top-k} for a subset of the Survey Response Generation Methods, more advanced decoding strategies~\citep[][inter alia]{zhang_sled_2024, garces_arias_decoding_2025} are also out of the scope of this paper.

We note that treating human survey responses as ground truth ignores biases in human survey response~\citep{groves_total_2010}. Future research should therefore consider evaluations of \textit{in-silico} survey responses that can be performed without this assumption.

\section*{Ethical Considerations}

Survey Response Generation Methods are frequently under-reported or insufficiently described in existing research, which poses challenges for reproducibility and transparency. In-silico surveys represent an emerging and active area of inquiry within both NLP and survey methodology. These approaches hold considerable promise for advancing survey research, for example to pre-test survey instruments or to impute missing data.

However, uncritical applications---particularly those aimed at directly predicting survey outcomes---carry the risk of distorting public opinion, with a heightened potential to misrepresent the perspectives of marginalized populations. Furthermore, unresolved epistemological questions persist regarding the extent to which simulated survey responses can meaningfully inform our understanding of the populations they aim to represent. Finally, issues of inferential privacy warrant careful consideration, as individuals may not consent to the simulation of their responses, particularly in cases where they have deliberately chosen not to participate in surveys.

\section*{Acknowledgments}
We would like to thank Marlene Lutz, Tobias Schumacher, Indira Sen, Florian Lemmerich, Florian Keusch, Frauke Kreuter, and the members of the FK\textsuperscript{2}RG research meeting for their helpful feedback on earlier versions of this project.

\newpage

\bibliography{SurveyLLMs}

@article{argyle_out_2023,
	title = {Out of {One}, {Many}: {Using} {Language} {Models} to {Simulate} {Human} {Samples}},
	volume = {31},
	issn = {1047-1987, 1476-4989},
	shorttitle = {Out of {One}, {Many}},
	url = {https://www.cambridge.org/core/product/identifier/S1047198723000025/type/journal_article},
	doi = {10.1017/pan.2023.2},
	language = {en},
	number = {3},
	urldate = {2024-01-17},
	journal = {Political Analysis},
	author = {Argyle, Lisa P. and Busby, Ethan C. and Fulda, Nancy and Gubler, Joshua R. and Rytting, Christopher and Wingate, David},
	month = jul,
	year = {2023},
	pages = {337--351},
}

@article{motoki_more_2023,
	title = {More human than human: measuring {ChatGPT} political bias},
	volume = {198},
	issn = {0048-5829, 1573-7101},
	shorttitle = {More human than human},
	url = {https://link.springer.com/10.1007/s11127-023-01097-2},
	doi = {10.1007/s11127-023-01097-2},
	language = {en},
	urldate = {2024-02-01},
	journal = {Public Choice},
	author = {Motoki, Fabio and Pinho Neto, Valdemar and Rodrigues, Victor},
	month = aug,
	year = {2023},
	pages = {3--23},
}

@article{groves_total_2010,
	title = {Total {Survey} {Error}: {Past}, {Present}, and {Future}},
	volume = {74},
	issn = {0033-362X, 1537-5331},
	shorttitle = {Total {Survey} {Error}},
	url = {https://academic.oup.com/poq/article-lookup/doi/10.1093/poq/nfq065},
	doi = {10.1093/poq/nfq065},
	language = {en},
	number = {5},
	urldate = {2024-02-20},
	journal = {Public Opinion Quarterly},
	author = {Groves, R. M. and Lyberg, L.},
	month = jan,
	year = {2010},
	pages = {849--879},
}

@misc{park_generative_2024,
	title = {Generative {Agent} {Simulations} of 1,000 {People}},
	url = {https://arxiv.org/abs/2411.10109},
	language = {en},
	publisher = {arXiv},
	author = {Park, Joon Sung and Zou, Carolyn Q and Shaw, Aaron and Hill, Benjamin Mako and Cai, Carrie and Morris, Meredith Ringel and Willer, Robb and Liang, Percy and Bernstein, Michael S},
	month = nov,
	year = {2024},
}

@inproceedings{wright_llm_2024,
	address = {Miami, Florida, USA},
	title = {{LLM} {Tropes}: {Revealing} {Fine}-{Grained} {Values} and {Opinions} in {Large} {Language} {Models}},
	shorttitle = {{LLM} {Tropes}},
	url = {https://aclanthology.org/2024.findings-emnlp.995},
	doi = {10.18653/v1/2024.findings-emnlp.995},
	language = {en},
	urldate = {2024-12-11},
	booktitle = {Findings of the {Association} for {Computational} {Linguistics}: {EMNLP} 2024},
	publisher = {Association for Computational Linguistics},
	author = {Wright, Dustin and Arora, Arnav and Borenstein, Nadav and Yadav, Srishti and Belongie, Serge and Augenstein, Isabelle},
	year = {2024},
	pages = {17085--17112},
}

@misc{suh_language_2025,
	title = {Language {Model} {Fine}-{Tuning} on {Scaled} {Survey} {Data} for {Predicting} {Distributions} of {Public} {Opinions}},
	url = {http://arxiv.org/abs/2502.16761},
	doi = {10.48550/arXiv.2502.16761},
	language = {en},
	urldate = {2025-02-27},
	publisher = {arXiv},
	author = {Suh, Joseph and Jahanparast, Erfan and Moon, Suhong and Kang, Minwoo and Chang, Serina},
	month = feb,
	year = {2025},
	note = {arXiv:2502.16761 [cs]},
}

@misc{sprague_cot_2024,
	title = {To {CoT} or not to {CoT}? {Chain}-of-thought helps mainly on math and symbolic reasoning},
	shorttitle = {To {CoT} or not to {CoT}?},
	url = {http://arxiv.org/abs/2409.12183},
	doi = {10.48550/arXiv.2409.12183},
	language = {en},
	urldate = {2025-03-14},
	publisher = {arXiv},
	author = {Sprague, Zayne and Yin, Fangcong and Rodriguez, Juan Diego and Jiang, Dongwei and Wadhwa, Manya and Singhal, Prasann and Zhao, Xinyu and Ye, Xi and Mahowald, Kyle and Durrett, Greg},
	month = oct,
	year = {2024},
	note = {arXiv:2409.12183 [cs]},
}

@inproceedings{kwon_efficient_2023,
	address = {Koblenz Germany},
	title = {Efficient {Memory} {Management} for {Large} {Language} {Model} {Serving} with {PagedAttention}},
	isbn = {979-8-4007-0229-7},
	shorttitle = {vllm},
	url = {https://dl.acm.org/doi/10.1145/3600006.3613165},
	doi = {10.1145/3600006.3613165},
	language = {en},
	urldate = {2025-06-20},
	booktitle = {Proceedings of the 29th {Symposium} on {Operating} {Systems} {Principles}},
	publisher = {ACM},
	author = {Kwon, Woosuk and Li, Zhuohan and Zhuang, Siyuan and Sheng, Ying and Zheng, Lianmin and Yu, Cody Hao and Gonzalez, Joseph and Zhang, Hao and Stoica, Ion},
	month = oct,
	year = {2023},
	pages = {611--626},
}

@misc{dong_xgrammar_2025,
	title = {{XGrammar}: {Flexible} and {Efficient} {Structured} {Generation} {Engine} for {Large} {Language} {Models}},
	shorttitle = {{XGrammar}},
	url = {http://arxiv.org/abs/2411.15100},
	doi = {10.48550/arXiv.2411.15100},
	urldate = {2025-06-20},
	publisher = {arXiv},
	author = {Dong, Yixin and Ruan, Charlie F. and Cai, Yaxing and Lai, Ruihang and Xu, Ziyi and Zhao, Yilong and Chen, Tianqi},
	month = may,
	year = {2025},
	note = {arXiv:2411.15100 [cs]},
}

@article{von_der_heyde_vox_2025,
	title = {Vox {Populi}, {Vox} {AI}? {Using} {Large} {Language} {Models} to {Estimate} {German} {Vote} {Choice}},
	copyright = {https://journals.sagepub.com/page/policies/text-and-data-mining-license},
	issn = {0894-4393, 1552-8286},
	shorttitle = {Vox {Populi}, {Vox} {AI}?},
	url = {https://journals.sagepub.com/doi/10.1177/08944393251337014},
	doi = {10.1177/08944393251337014},
	language = {en},
	urldate = {2025-07-20},
	journal = {Social Science Computer Review},
	publisher = {SAGE Publications},
	author = {Von Der Heyde, Leah and Haensch, Anna-Carolina and Wenz, Alexander},
	month = apr,
	year = {2025},
}

@article{boelaert_machine_2025,
	title = {Machine {Bias}. {How} {Do} {Generative} {Language} {Models} {Answer} {Opinion} {Polls}?},
	volume = {54},
	issn = {0049-1241, 1552-8294},
	shorttitle = {Machine {Bias}. {How} {Do} {Generative} {Language} {Models} {Answer} {Opinion} {Polls}?},
	url = {https://journals.sagepub.com/doi/10.1177/00491241251330582},
	doi = {10.1177/00491241251330582},
	language = {en},
	number = {3},
	urldate = {2025-08-12},
	journal = {Sociological Methods \& Research},
	author = {Boelaert, Julien and Coavoux, Samuel and Ollion, Étienne and Petev, Ivaylo and Präg, Patrick},
	month = aug,
	year = {2025},
	pages = {1156--1196},
}

@misc{gles_gles_2025,
	title = {{GLES} 2025 {Post}-{Election} {Cross} {Section}},
	url = {https://search.gesis.org/research_data/ZA10100?doi=10.4232/5.ZA10100.1.0.0},
	doi = {10.4232/5.ZA10100.1.0.0},
	urldate = {2025-09-10},
	publisher = {GESIS Data Archive},
	author = {{GLES}},
	year = {2025},
}

@misc{gles_gles_2017,
	title = {{GLES} 2017 {Post}-{Election} {Cross} {Section}},
	copyright = {Alle im GESIS DBK veröffentlichten Metadaten sind frei verfügbar unter den Creative Commons CC0 1.0 Universal Public Domain Dedication. GESIS bittet jedoch darum, dass Sie alle Metadatenquellen anerkennen und sie nennen, etwa die Datengeber oder jeglichen Aggregator, inklusive GESIS selbst. Für weitere Informationen siehe https://dbk.gesis.org/dbksearch/guidelines.asp?db=d, All metadata from GESIS DBK are available free of restriction under the Creative Commons CC0 1.0 Universal Public Domain Dedication. However, GESIS requests that you actively acknowledge and give attribution to all metadata sources, such as the data providers and any data aggregators, including GESIS. For further information see https://dbk.gesis.org/dbksearch/guidelines.asp},
	url = {https://search.gesis.org/research_data/ZA6801?doi=10.4232/1.13235},
	doi = {10.4232/1.13235},
	language = {de},
	urldate = {2025-09-10},
	publisher = {GESIS Data Archive},
	author = {{GLES}},
	collaborator = {Roßteutscher, Sigrid and Schmitt-Beck, Rüdiger and Schoen, Harald and Weßels, Bernhard and Wolf, Christof and Wagner, Aiko and Blumenberg, Manuela and Förster, André and Giebler, Heiko and Melcher, Reinhold and Roßmann, Joss and Blinzler, Katharina and Jungmann, Nils and Kratz, Agatha and Kratz, Sophia and {Kantar Public Germany}},
	year = {2017},
}

@misc{rupprecht_prompt_2025,
	title = {Prompt {Perturbations} {Reveal} {Human}-{Like} {Biases} in {LLM} {Survey} {Responses}},
	url = {http://arxiv.org/abs/2507.07188},
	doi = {10.48550/arXiv.2507.07188},
	language = {en},
	urldate = {2025-09-11},
	publisher = {arXiv},
	author = {Rupprecht, Jens and Ahnert, Georg and Strohmaier, Markus},
	month = jul,
	year = {2025},
	note = {arXiv:2507.07188 [cs]},
}

@misc{myrzakhan_open-llm-leaderboard_2024,
	title = {Open-{LLM}-{Leaderboard}: {From} {Multi}-choice to {Open}-style {Questions} for {LLMs} {Evaluation}, {Benchmark}, and {Arena}},
	shorttitle = {Open-{LLM}-{Leaderboard}},
	url = {http://arxiv.org/abs/2406.07545},
	doi = {10.48550/arXiv.2406.07545},
	language = {en},
	urldate = {2025-09-26},
	publisher = {arXiv},
	author = {Myrzakhan, Aidar and Bsharat, Sondos Mahmoud and Shen, Zhiqiang},
	month = jun,
	year = {2024},
	note = {arXiv:2406.07545 [cs]},
}

@inproceedings{zhang_sled_2024,
	title = {{SLED}: {Self} {Logits} {Evolution} {Decoding} for {Improving} {Factuality} in {Large} {Language} {Models}},
	volume = {37},
	url = {https://proceedings.neurips.cc/paper_files/paper/2024/file/0939f13ffce3ff487509d902ddba4571-Paper-Conference.pdf},
	booktitle = {Advances in {Neural} {Information} {Processing} {Systems}},
	publisher = {Curran Associates, Inc.},
	author = {Zhang, Jianyi and Juan, Da-Cheng and Rashtchian, Cyrus and Ferng, Chun-Sung and Jiang, Heinrich and Chen, Yiran},
	editor = {Globerson, A. and Mackey, L. and Belgrave, D. and Fan, A. and Paquet, U. and Tomczak, J. and Zhang, C.},
	year = {2024},
	pages = {5188--5209},
}

@misc{cummins_threat_2025,
	title = {The threat of analytic flexibility in using large language models to simulate human data: {A} call to attention},
	shorttitle = {The threat of analytic flexibility in using large language models to simulate human data},
	url = {http://arxiv.org/abs/2509.13397},
	doi = {10.48550/arXiv.2509.13397},
	urldate = {2025-09-28},
	publisher = {arXiv},
	author = {Cummins, Jamie},
	month = sep,
	year = {2025},
	note = {arXiv:2509.13397 [cs]},
}

@article{szekely_measuring_2007,
	title = {Measuring and testing dependence by correlation of distances},
	volume = {35},
	issn = {0090-5364},
	url = {https://projecteuclid.org/journals/annals-of-statistics/volume-35/issue-6/Measuring-and-testing-dependence-by-correlation-of-distances/10.1214/009053607000000505.full},
	doi = {10.1214/009053607000000505},
	language = {en},
	number = {6},
	urldate = {2025-10-02},
	journal = {The Annals of Statistics},
	author = {Székely, Gábor J. and Rizzo, Maria L. and Bakirov, Nail K.},
	month = dec,
	year = {2007},
}

@misc{holtdirk_learning_2025,
	title = {Learning from {Convenience} {Samples}: {A} {Case} {Study} on {Fine}-{Tuning} {LLMs} for {Survey} {Non}-response in the {German} {Longitudinal} {Election} {Study}},
	shorttitle = {Learning from {Convenience} {Samples}},
	url = {http://arxiv.org/abs/2509.25063},
	doi = {10.48550/arXiv.2509.25063},
	language = {en},
	urldate = {2025-10-03},
	publisher = {arXiv},
	author = {Holtdirk, Tobias and Assenmacher, Dennis and Bleier, Arnim and Wagner, Claudia},
	month = sep,
	year = {2025},
	note = {arXiv:2509.25063 [cs]},
}

@inproceedings{balepur_which_2025,
	address = {Vienna, Austria},
	title = {Which of {These} {Best} {Describes} {Multiple} {Choice} {Evaluation} with {LLMs}? {A}) {Forced} {B}) {Flawed} {C}) {Fixable} {D}) {All} of the {Above}},
	isbn = {979-8-89176-251-0},
	shorttitle = {Which of {These} {Best} {Describes} {Multiple} {Choice} {Evaluation} with {LLMs}?},
	url = {https://aclanthology.org/2025.acl-long.169/},
	doi = {10.18653/v1/2025.acl-long.169},
	urldate = {2025-10-03},
	booktitle = {Proceedings of the 63rd {Annual} {Meeting} of the {Association} for {Computational} {Linguistics} ({Volume} 1: {Long} {Papers})},
	publisher = {Association for Computational Linguistics},
	author = {Balepur, Nishant and Rudinger, Rachel and Boyd-Graber, Jordan Lee},
	editor = {Che, Wanxiang and Nabende, Joyce and Shutova, Ekaterina and Pilehvar, Mohammad Taher},
	month = jul,
	year = {2025},
	pages = {3394--3418},
}

@inproceedings{sorensen_position_2024,
	address = {Vienna, Austria},
	series = {{ICML}'24},
	title = {Position: a roadmap to pluralistic alignment},
	volume = {235},
	shorttitle = {Position},
	url = {https://dl.acm.org/doi/abs/10.5555/3692070.3693952},
	urldate = {2025-10-05},
	booktitle = {Proceedings of the 41st {International} {Conference} on {Machine} {Learning}},
	publisher = {JMLR.org},
	author = {Sorensen, Taylor and Moore, Jared and Fisher, Jillian and Gordon, Mitchell and Mireshghallah, Niloofar and Rytting, Christopher Michael and Ye, Andre and Jiang, Liwei and Lu, Ximing and Dziri, Nouha and Althoff, Tim and Choi, Yejin},
	month = jul,
	year = {2024},
	pages = {46280--46302},
}

@misc{anes_2016_2016,
	title = {2016 {Time} {Series} {Study}},
	url = {https://electionstudies.org/data-center/2016-time-series-study/},
	language = {en-US},
	urldate = {2025-10-07},
	author = {{ANES}},
	year = {2016},
}

@misc{atp_american_2021,
	title = {The {American} {Trends} {Panel}},
	url = {https://www.pewresearch.org/the-american-trends-panel/},
	language = {en-US},
	urldate = {2025-10-07},
	author = {{ATP}},
	year = {2021},
}

@misc{qwen_team_qwen3_2025,
	title = {Qwen3 {Technical} {Report}},
	url = {http://arxiv.org/abs/2505.09388},
	doi = {10.48550/arXiv.2505.09388},
	language = {en},
	urldate = {2025-10-07},
	publisher = {arXiv},
	author = {{Qwen Team}},
	month = may,
	year = {2025},
	note = {arXiv:2505.09388 [cs]},
}

@misc{llama_team_llama_2024,
	title = {The {Llama} 3 {Herd} of {Models}},
	url = {http://arxiv.org/abs/2407.21783},
	doi = {10.48550/arXiv.2407.21783},
	language = {en},
	urldate = {2025-10-07},
	publisher = {arXiv},
	author = {{Llama Team}},
	month = nov,
	year = {2024},
	note = {arXiv:2407.21783 [cs]},
}

@misc{olmo_team_2_2025,
	title = {2 {OLMo} 2 {Furious}},
	url = {http://arxiv.org/abs/2501.00656},
	doi = {10.48550/arXiv.2501.00656},
	language = {en},
	urldate = {2025-10-07},
	publisher = {arXiv},
	author = {{OLMo Team}},
	month = jan,
	year = {2025},
	note = {arXiv:2501.00656 [cs]},
}

@inproceedings{garces_arias_decoding_2025,
	address = {Abu Dhabi, UAE},
	title = {Decoding {Decoded}: {Understanding} {Hyperparameter} {Effects} in {Open}-{Ended} {Text} {Generation}},
	shorttitle = {Decoding {Decoded}},
	url = {https://aclanthology.org/2025.coling-main.668/},
	urldate = {2025-10-13},
	booktitle = {Proceedings of the 31st {International} {Conference} on {Computational} {Linguistics}},
	publisher = {Association for Computational Linguistics},
	author = {Garces Arias, Esteban and Li, Meimingwei and Heumann, Christian and Assenmacher, Matthias},
	editor = {Rambow, Owen and Wanner, Leo and Apidianaki, Marianna and Al-Khalifa, Hend and Eugenio, Barbara Di and Schockaert, Steven},
	month = jan,
	year = {2025},
	pages = {9992--10020},
}

@inproceedings{chen_seeing_2024,
	address = {Miami, Florida, USA},
	title = {“{Seeing} the {Big} through the {Small}”: {Can} {LLMs} {Approximate} {Human} {Judgment} {Distributions} on {NLI} from a {Few} {Explanations}?},
	shorttitle = {“{Seeing} the {Big} through the {Small}”},
	url = {https://aclanthology.org/2024.findings-emnlp.842},
	doi = {10.18653/v1/2024.findings-emnlp.842},
	language = {en},
	urldate = {2025-10-13},
	booktitle = {Findings of the {Association} for {Computational} {Linguistics}: {EMNLP} 2024},
	publisher = {Association for Computational Linguistics},
	author = {Chen, Beiduo and Wang, Xinpeng and Peng, Siyao and Litschko, Robert and Korhonen, Anna and Plank, Barbara},
	year = {2024},
	pages = {14396--14419},
}

@inproceedings{dominguez-olmedo_questioning_2024,
	title = {Questioning the {Survey} {Responses} of {Large} {Language} {Models}},
	volume = {37},
	url = {https://proceedings.neurips.cc/paper_files/paper/2024/hash/515c62809e0a29729d7eec26e2916fc0-Abstract-Conference.html},
	language = {en},
	urldate = {2025-10-13},
	booktitle = {Advances in {Neural} {Information} {Processing} {Systems}},
	author = {Dominguez-Olmedo, Ricardo and Hardt, Moritz and Mendler-Dünner, Celestine},
	month = dec,
	year = {2024},
	pages = {45850--45878},
}

@misc{hartmann_political_2023,
	address = {Rochester, NY},
	type = {{SSRN} {Scholarly} {Paper}},
	title = {The political ideology of conversational {AI}: {Converging} evidence on {ChatGPT}’s pro-environmental, left-libertarian orientation},
	shorttitle = {The political ideology of conversational {AI}},
	url = {https://papers.ssrn.com/abstract=4316084},
	doi = {10.2139/ssrn.4316084},
	language = {en},
	urldate = {2025-10-13},
	publisher = {Social Science Research Network},
	author = {Hartmann, Jochen and Schwenzow, Jasper and Witte, Maximilian},
	month = jan,
	year = {2023},
}

@inproceedings{ma_potential_2024,
	address = {Miami, Florida, USA},
	title = {The {Potential} and {Challenges} of {Evaluating} {Attitudes}, {Opinions}, and {Values} in {Large} {Language} {Models}},
	url = {https://aclanthology.org/2024.findings-emnlp.513/},
	doi = {10.18653/v1/2024.findings-emnlp.513},
	urldate = {2025-10-13},
	booktitle = {Findings of the {Association} for {Computational} {Linguistics}: {EMNLP} 2024},
	publisher = {Association for Computational Linguistics},
	author = {Ma, Bolei and Wang, Xinpeng and Hu, Tiancheng and Haensch, Anna-Carolina and Hedderich, Michael A. and Plank, Barbara and Kreuter, Frauke},
	editor = {Al-Onaizan, Yaser and Bansal, Mohit and Chen, Yun-Nung},
	month = nov,
	year = {2024},
	pages = {8783--8805},
}

@article{mcilroy-young_order-independence_2024,
	title = {Order-{Independence} {Without} {Fine} {Tuning}},
	volume = {37},
	url = {https://proceedings.neurips.cc/paper_files/paper/2024/hash/85529bc995777a74072ef63c05bedd30-Abstract-Conference.html},
	language = {en},
	urldate = {2025-10-13},
	journal = {Advances in Neural Information Processing Systems},
	author = {McIlroy-Young, Reid and Brown, Katrina and Olson, Conlan and Zhang, Linjun and Dwork, Cynthia},
	month = dec,
	year = {2024},
	pages = {72818--72839},
}

@inproceedings{meister_benchmarking_2025,
	address = {Albuquerque, New Mexico},
	title = {Benchmarking {Distributional} {Alignment} of {Large} {Language} {Models}},
	isbn = {979-8-89176-189-6},
	url = {https://aclanthology.org/2025.naacl-long.2/},
	doi = {10.18653/v1/2025.naacl-long.2},
	urldate = {2025-10-13},
	booktitle = {Proceedings of the 2025 {Conference} of the {Nations} of the {Americas} {Chapter} of the {Association} for {Computational} {Linguistics}: {Human} {Language} {Technologies} ({Volume} 1: {Long} {Papers})},
	publisher = {Association for Computational Linguistics},
	author = {Meister, Nicole and Guestrin, Carlos and Hashimoto, Tatsunori},
	editor = {Chiruzzo, Luis and Ritter, Alan and Wang, Lu},
	month = apr,
	year = {2025},
	pages = {24--49},
}

@inproceedings{rottger_political_2024,
	address = {Bangkok, Thailand},
	title = {Political {Compass} or {Spinning} {Arrow}? {Towards} {More} {Meaningful} {Evaluations} for {Values} and {Opinions} in {Large} {Language} {Models}},
	shorttitle = {Political {Compass} or {Spinning} {Arrow}?},
	url = {https://aclanthology.org/2024.acl-long.816/},
	doi = {10.18653/v1/2024.acl-long.816},
	urldate = {2025-10-13},
	booktitle = {Proceedings of the 62nd {Annual} {Meeting} of the {Association} for {Computational} {Linguistics} ({Volume} 1: {Long} {Papers})},
	publisher = {Association for Computational Linguistics},
	author = {Röttger, Paul and Hofmann, Valentin and Pyatkin, Valentina and Hinck, Musashi and Kirk, Hannah and Schuetze, Hinrich and Hovy, Dirk},
	editor = {Ku, Lun-Wei and Martins, Andre and Srikumar, Vivek},
	month = aug,
	year = {2024},
	pages = {15295--15311},
}

@inproceedings{santurkar_whose_2023,
	title = {Whose {Opinions} {Do} {Language} {Models} {Reflect}?},
	issn = {2640-3498},
	url = {https://proceedings.mlr.press/v202/santurkar23a.html},
	language = {en},
	urldate = {2025-10-13},
	booktitle = {Proceedings of the 40th {International} {Conference} on {Machine} {Learning}},
	publisher = {PMLR},
	author = {Santurkar, Shibani and Durmus, Esin and Ladhak, Faisal and Lee, Cinoo and Liang, Percy and Hashimoto, Tatsunori},
	month = jul,
	year = {2023},
	pages = {29971--30004},
}

@inproceedings{tam_let_2024,
	address = {Miami, Florida, US},
	title = {Let {Me} {Speak} {Freely}? {A} {Study} {On} {The} {Impact} {Of} {Format} {Restrictions} {On} {Large} {Language} {Model} {Performance}.},
	shorttitle = {Let {Me} {Speak} {Freely}?},
	url = {https://aclanthology.org/2024.emnlp-industry.91/},
	doi = {10.18653/v1/2024.emnlp-industry.91},
	urldate = {2025-10-13},
	booktitle = {Proceedings of the 2024 {Conference} on {Empirical} {Methods} in {Natural} {Language} {Processing}: {Industry} {Track}},
	publisher = {Association for Computational Linguistics},
	author = {Tam, Zhi Rui and Wu, Cheng-Kuang and Tsai, Yi-Lin and Lin, Chieh-Yen and Lee, Hung-yi and Chen, Yun-Nung},
	editor = {Dernoncourt, Franck and Preoţiuc-Pietro, Daniel and Shimorina, Anastasia},
	month = nov,
	year = {2024},
	pages = {1218--1236},
}

@article{tjuatja_llms_2024,
	title = {Do {LLMs} {Exhibit} {Human}-like {Response} {Biases}? {A} {Case} {Study} in {Survey} {Design}},
	volume = {12},
	issn = {2307-387X},
	shorttitle = {Do {LLMs} {Exhibit} {Human}-like {Response} {Biases}?},
	url = {https://doi.org/10.1162/tacl_a_00685},
	doi = {10.1162/tacl_a_00685},
	urldate = {2025-10-13},
	journal = {Transactions of the Association for Computational Linguistics},
	author = {Tjuatja, Lindia and Chen, Valerie and Wu, Tongshuang and Talwalkwar, Ameet and Neubig, Graham},
	month = sep,
	year = {2024},
	pages = {1011--1026},
}

@inproceedings{wang_my_2024,
	address = {Bangkok, Thailand},
	title = {“{My} {Answer} is {C}”: {First}-{Token} {Probabilities} {Do} {Not} {Match} {Text} {Answers} in {Instruction}-{Tuned} {Language} {Models}},
	shorttitle = {“{My} {Answer} is {C}”},
	url = {https://aclanthology.org/2024.findings-acl.441/},
	doi = {10.18653/v1/2024.findings-acl.441},
	urldate = {2025-10-13},
	booktitle = {Findings of the {Association} for {Computational} {Linguistics}: {ACL} 2024},
	publisher = {Association for Computational Linguistics},
	author = {Wang, Xinpeng and Ma, Bolei and Hu, Chengzhi and Weber-Genzel, Leon and Röttger, Paul and Kreuter, Frauke and Hovy, Dirk and Plank, Barbara},
	editor = {Ku, Lun-Wei and Martins, Andre and Srikumar, Vivek},
	month = aug,
	year = {2024},
	pages = {7407--7416},
}

@misc{hu_simbench_2025,
	title = {{SimBench}: {Benchmarking} the {Ability} of {Large} {Language} {Models} to {Simulate} {Human} {Behaviors}},
	shorttitle = {{SimBench}},
	url = {http://arxiv.org/abs/2510.17516},
	doi = {10.48550/arXiv.2510.17516},
	urldate = {2025-11-14},
	publisher = {arXiv},
	author = {Hu, Tiancheng and Baumann, Joachim and Lupo, Lorenzo and Collier, Nigel and Hovy, Dirk and Röttger, Paul},
	month = oct,
	year = {2025},
	note = {arXiv:2510.17516 [cs]},
}

@inproceedings{licht_measuring_2025,
	address = {Suzhou, China},
	title = {Measuring scalar constructs in social science with {LLMs}},
	isbn = {979-8-89176-332-6},
	url = {https://aclanthology.org/2025.emnlp-main.1635/},
	doi = {10.18653/v1/2025.emnlp-main.1635},
	urldate = {2025-12-03},
	booktitle = {Proceedings of the 2025 {Conference} on {Empirical} {Methods} in {Natural} {Language} {Processing}},
	publisher = {Association for Computational Linguistics},
	author = {Licht, Hauke and Sarkar, Rupak and Wu, Patrick Y. and Goel, Pranav and Stoehr, Niklas and Ash, Elliott and Hoyle, Alexander Miserlis},
	editor = {Christodoulopoulos, Christos and Chakraborty, Tanmoy and Rose, Carolyn and Peng, Violet},
	month = nov,
	year = {2025},
	pages = {32132--32159},
}

@inproceedings{lutz_prompt_2025,
	address = {Suzhou, China},
	title = {The {Prompt} {Makes} the {Person}(a): {A} {Systematic} {Evaluation} of {Sociodemographic} {Persona} {Prompting} for {Large} {Language} {Models}},
	isbn = {979-8-89176-335-7},
	shorttitle = {The {Prompt} {Makes} the {Person}(a)},
	url = {https://aclanthology.org/2025.findings-emnlp.1261/},
	doi = {10.18653/v1/2025.findings-emnlp.1261},
	urldate = {2026-01-30},
	booktitle = {Findings of the {Association} for {Computational} {Linguistics}: {EMNLP} 2025},
	publisher = {Association for Computational Linguistics},
	author = {Lutz, Marlene and Sen, Indira and Ahnert, Georg and Rogers, Elisa and Strohmaier, Markus},
	editor = {Christodoulopoulos, Christos and Chakraborty, Tanmoy and Rose, Carolyn and Peng, Violet},
	month = nov,
	year = {2025},
	pages = {23212--23237},
}

@inproceedings{kreutner_qstn_2026,
	address = {Rabat, Marocco},
	title = {{QSTN}: {A} {Modular} {Framework} for {Robust} {Questionnaire} {Inference} with {Large} {Language} {Models}},
	isbn = {979-8-89176-382-1},
	shorttitle = {{QSTN}},
	url = {https://aclanthology.org/2026.eacl-demo.37/},
	doi = {10.18653/v1/2026.eacl-demo.37},
	urldate = {2026-04-17},
	booktitle = {Proceedings of the 19th {Conference} of the {European} {Chapter} of the {Association} for {Computational} {Linguistics} ({Volume} 3: {System} {Demonstrations})},
	publisher = {Association for Computational Linguistics},
	author = {Kreutner, Maximilian and Rupprecht, Jens and Ahnert, Georg and Salem, Ahmed and Strohmaier, Markus},
	editor = {Croce, Danilo and Leidner, Jochen and Moosavi, Nafise Sadat},
	month = mar,
	year = {2026},
	pages = {537--549},
}

\appendix

\begin{figure}[t!]
    \centering
    \includegraphics[width=\linewidth]{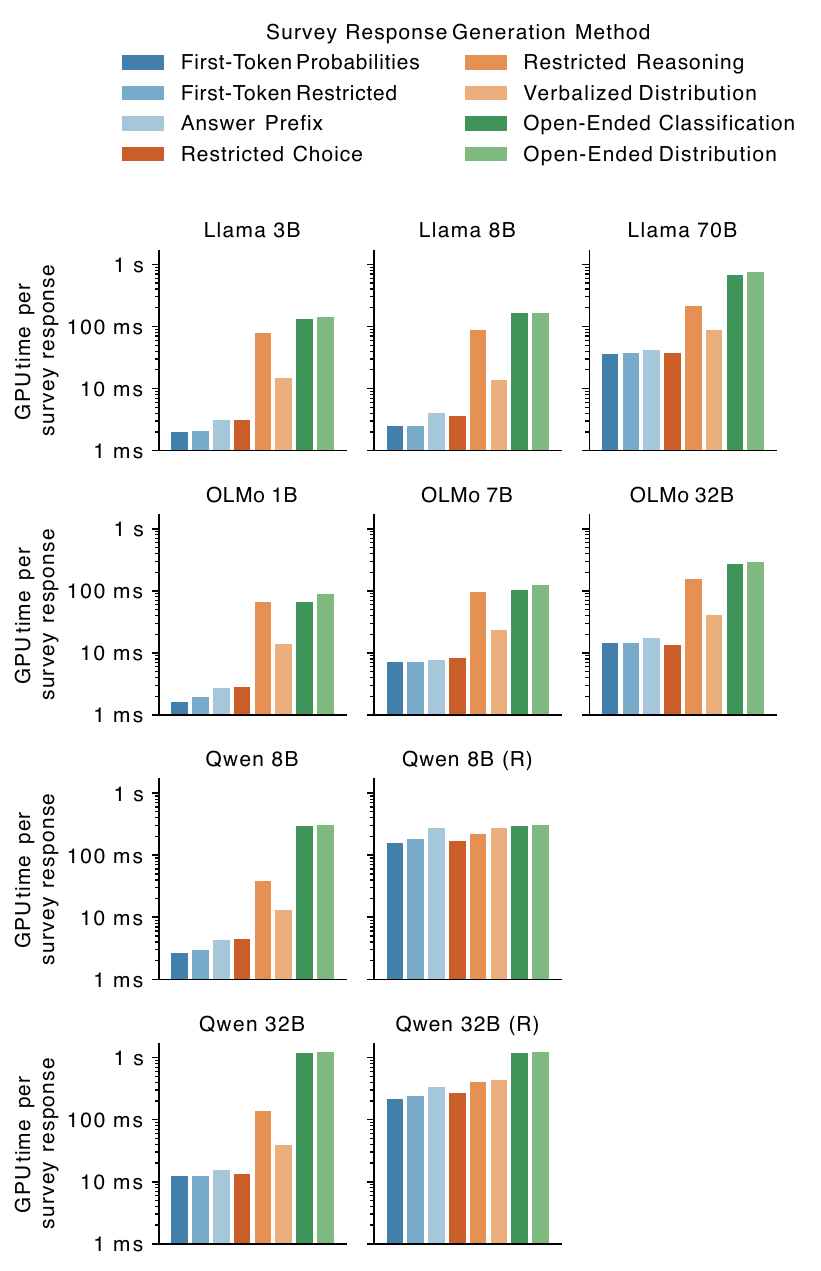}
    \caption{\textbf{Mean GPU Time for A Single Survey Response.} We run all models using vllm on 2 NVIDIA H100 GPUs (tensor-parallel). We report GPU time instead of token count to accommodate for optimizations such as automatic prefix caching, but also for the overhead that is created by restricting the vocabulary of an LLM with structured outputs. Considering the log-scale y-axis, Open Generation Methods, larger models, and in particular reasoning models require \textbf{orders of magnitude more GPU time} than Token Probability-Based Methods, or the Restricted Choice Method.}
    \label{app_fig:GPU_time}
\end{figure}

\begin{table}[b!]
    \centering
    \small
    \begin{tblr}{
      colsep=5pt,
      rowsep=0.5pt,
      column{1}={2-Z}{font=\bfseries},
      column{2}={2-Z}{font=\ttfamily},
      row{1}={font=\bfseries},
      vline{2} = {-}{},
      hline{2} = {-}{},
      hborder{2}={belowspace=3pt},
    }
    Short Name     & Huggingface Model ID \\
    Llama\,3B      & \href{https://huggingface.co/meta-llama/Llama-3.2-3B-Instruct}{meta-llama/Llama-3.2-3B-Instruct} \\
    Llama\,8B      & \href{https://huggingface.co/meta-llama/Llama-3.1-8B-Instruct}{meta-llama/Llama-3.1-8B-Instruct} \\
    Llama\,70B     & \href{https://huggingface.co/meta-llama/Llama-3.3-70B-Instruct}{meta-llama/Llama-3.3-70B-Instruct} \\
    OLMo\,1B       & \href{https://huggingface.co/allenai/OLMo-2-0425-1B-Instruct}{allenai/OLMo-2-0425-1B-Instruct} \\
    OLMo\,7B       & \href{https://huggingface.co/allenai/OLMo-2-1124-7B-Instruct}{allenai/OLMo-2-1124-7B-Instruct} \\
    OLMo\,32B      & \href{https://huggingface.co/allenai/OLMo-2-0325-32B-Instruct}{allenai/OLMo-2-0325-32B-Instruct} \\
    Qwen\,8B       & \href{https://huggingface.co/Qwen/Qwen3-8B}{Qwen/Qwen3-8B} \\
    Qwen\,32B      & \href{https://huggingface.co/Qwen/Qwen3-32B}{Qwen/Qwen3-32B} \\
    Qwen\,8B\,(R)  & \href{https://huggingface.co/Qwen/Qwen3-8B}{Qwen/Qwen3-8B} with Reasoning \\
    Qwen\,32B\,(R) & \href{https://huggingface.co/Qwen/Qwen3-32B}{Qwen/Qwen3-32B} with Reasoning
    \end{tblr}
    \caption{\textbf{Language Models.} We evaluate all Survey Response Generation Methods on 10 open-weight LLMs.}
    \label{app_tab:language_models}
\end{table}

\section{Computational Details} \label{sec:computational_details}
We run all our experiments the 10 open-weight instruction tuned and reasoning models shown in Table~\ref{app_tab:language_models}.
For language model inference, we use \texttt{vllm} \cite{kwon_efficient_2023} version \texttt{0.10.1.1} and the \texttt{xgrammar} \cite{dong_xgrammar_2025} backend for inference with \texttt{structured outputs}. We use the \texttt{QSTN} framework~\citep{kreutner_qstn_2026} to facilitate prompt perturbations and the configuration of structured outputs. We ran all our experiments on 2 NVIDIA H100 GPUs (tensor-parallel). Our experiments for the ANES 2016 dataset had a total runtime of 88h, for the GLES 2017 dataset of 121h, for the GLES 2025 dataset of 297h, and for the ATP 2021 dataset of 108h. We ran our additional experiments on the impact of decoding hyper-parameters (see Figure~\ref{app_fig:temp_topk}) on 2 NVIDIA RTX PRO 6000 Blackwell GPUs with a total runtime of 29h. To save computational resources, we only generated open-ended output once for each simulation specification (model, seed, response scale, temperature) and then classified the same output separately for the Open-Ended Classification and for the Open-Ended Distribution Methods. Figure~\ref{app_fig:GPU_time} shows the average GPU Time spent to generate a single survey response with a given Survey Response Generation Method and LLM.

\section{Prompts} \label{sec:instruction_prompts}
Tables~\ref{tab:EN_system_prompts}--\ref{tab:persona_prompts} contain all system and user prompts that we used in our evaluations.

\section{Additional Results} \label{sec:additional_results}

\begin{figure*}[t!]
    \centering
    \includegraphics[width=\textwidth]{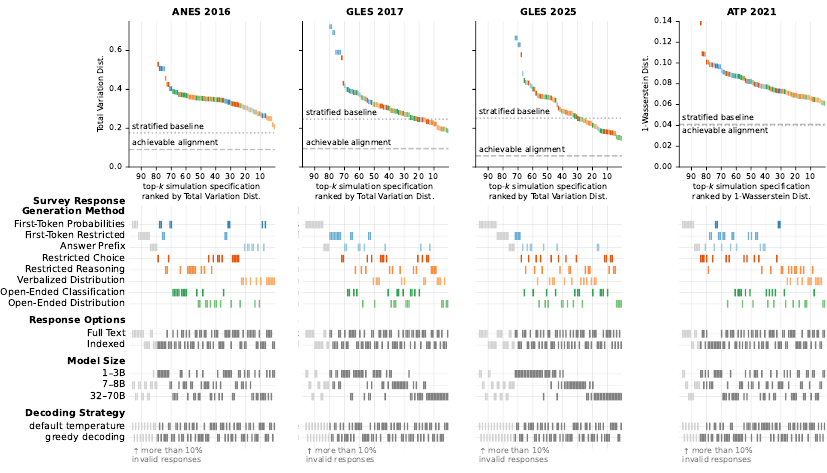}
    \caption{\textbf{Subpopulation-Level Alignment: Total Variation Distance/1-Wasserstein Distance.} For the ANES and GLES datasets, we use total variation distance to measure alignment on categorical response options. For the ATP dataset, we use 1-Wasserstein Distance to measure alignment on ordinal response options.
    \textbf{Top:} alignment metric (lower is better) for each aggregated simulation specification, mean across the respective runs. \textbf{Bottom:} simulation specification---Survey Response Generation Method, response option variant, model size, and decoding strategy---sorted by the respective alignment metric.
    Specifications that lead to more than 10\% invalid responses are excluded.
    }
    \label{app_fig:all_datasets_tv_specification}
\end{figure*}


\begin{figure*}[t!]
    \centering
    \includegraphics[width=\textwidth]{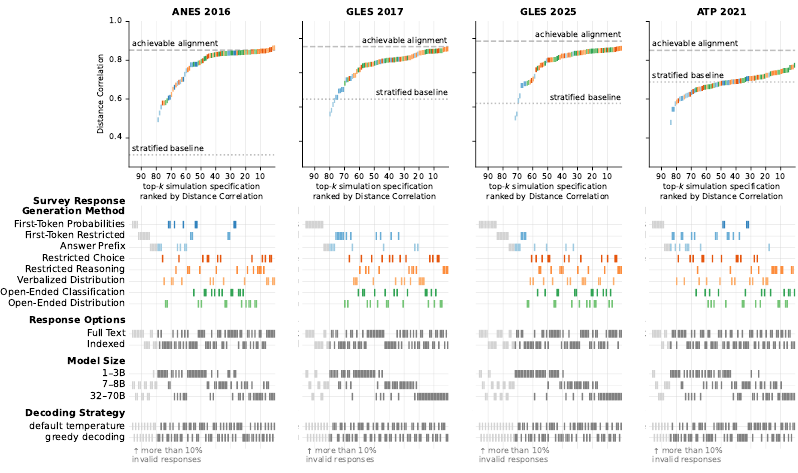}
    \caption{\textbf{Subpopulation-Level Alignment---Global Perspective: Distance Correlation.}
    \textbf{Top:} Distance correlation (higher is better) for each aggregated simulation specification, mean across the respective runs. \textbf{Bottom:} simulation specification---Survey Response Generation Method, response option variant, model size, and decoding strategy---sorted by distance correlation.
    Specifications that lead to more than 10\% invalid responses are excluded.
    }
    \label{app_fig:all_datasets_dcor_specification}
\end{figure*}

\begin{figure*}
    \begin{subfigure}{0.48\textwidth}
        \centering
        \includegraphics[width=\linewidth]{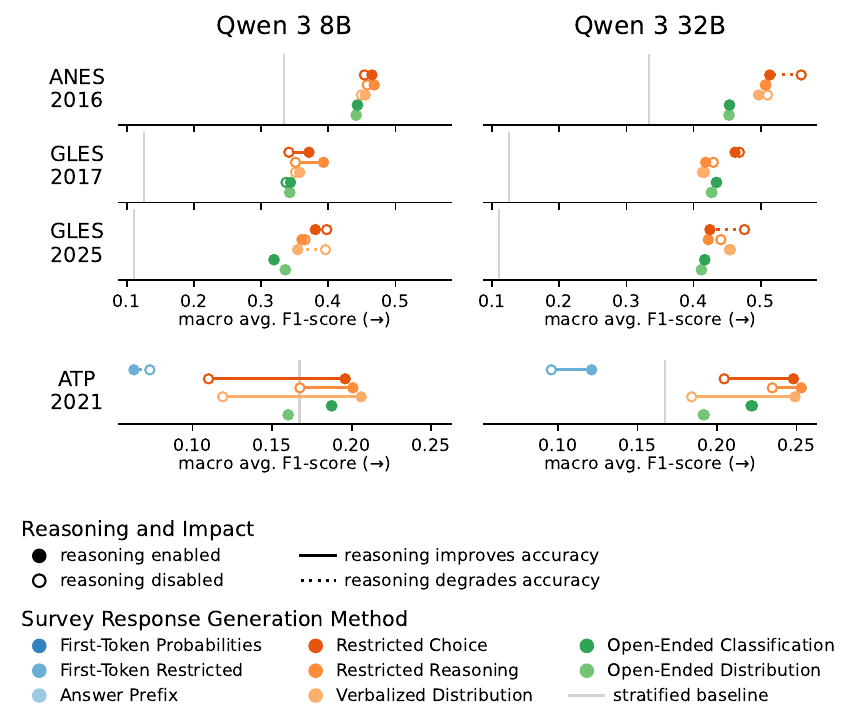}
        \caption{\textbf{Individual-Level Alignment.}}
        \label{app_fig:reasoning_models_acc}
    \end{subfigure}
    \begin{subfigure}{0.48\textwidth}
        \centering
        \includegraphics[width=\linewidth]{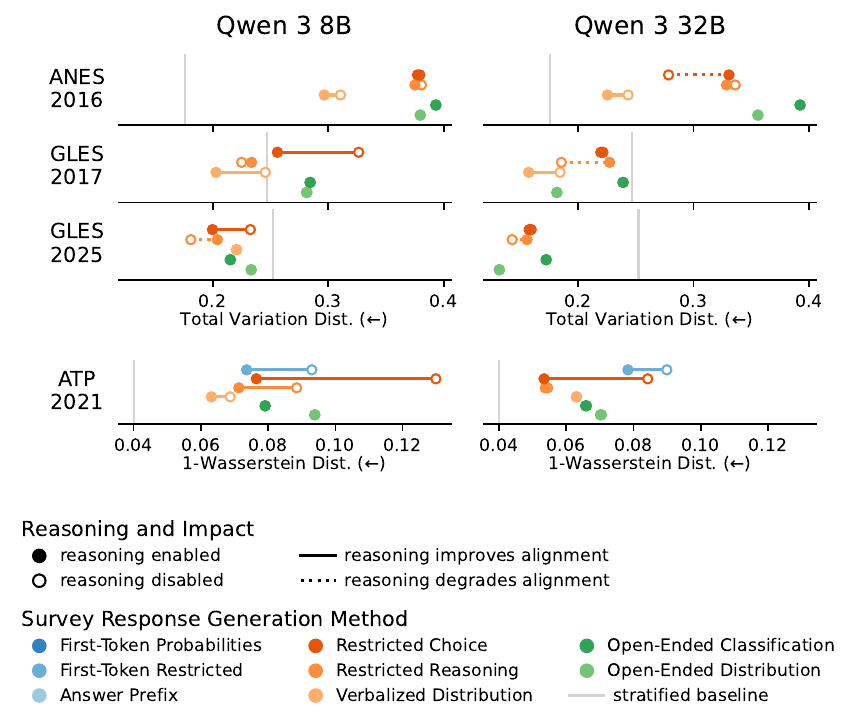}
        \caption{\textbf{Subpopulation-Level Alignment.}}
        \label{fig:reasoning_models_align}
    \end{subfigure}
    \caption{\textbf{Alignment With / Without Reasoning.} Mean results shown for default temperature, Full Text response option variants. Methods with more than 1\% invalid responses are excluded. \textbf{Reasoning does not consistently improve alignment.}}
    \label{fig:reasoning_models_both}
\end{figure*}

\begin{table}[ht!]
    \centering
    \footnotesize
    \begin{tblr}{
      colsep=3pt,
      colspec={Q[1.5mm, colsep=0pt] Q[l, font=\bfseries, colsep=4pt] | X[l] X[l] X[l] X[l]}, 
      row{1}={font=\bfseries},
      rowsep=2pt,
      width=\linewidth,
      hline{2,3,5,8}={black},
      hborder{2,3,5,8}={belowspace=3pt}
    }
& & {ANES\\2016} & {GLES\\2017} & {GLES\\2025} & {ATP\\2021} \\
\SetCell{bg=tab_blue} & Intercept & .450* & .088* & .305* & .204* \\
\SetCell{bg=tab_blue2} & First-Token Restricted & .060* & .138* & -.073 & -.029* \\
\SetCell{bg=tab_blue3} & Answer Prefix & .009 & .142* & -.039 & -.022 \\
\SetCell{bg=tab_orange} & Restricted Choice & \underline{.102*} & \textbf{.298*} & \textbf{.116*} & .006 \\
\SetCell{bg=tab_orange2} & Restricted Reasoning & \textbf{.110*} & \underline{.289*} & .055 & \textbf{.030*} \\
\SetCell{bg=tab_orange3} & Verbalized Distribution & .079* & .288* & .105 & .001 \\
\SetCell{bg=tab_green} & Open-Ended Classif. & .100* & .266* & .076 & \underline{.019} \\
\SetCell{bg=tab_green2} & Open-Ended Distrib. & .096* & .263* & .061 & .010 \\
    \end{tblr}
    \caption{\textbf{Regression Coefficients for Individual-Level Accuracy $(\uparrow)$.}
    OLS regression for each dataset with accuracy $(\uparrow)$ in each simulation specification as the dependent variable. We use Survey Response Generation Method, response option scale, and LLM as independent variables. We show coefficients for the Survey Response Generation Methods (Reference: First-Token Probabilities~{\color{tab_blue}$\blacksquare$}). For macro avg.\ F1-score as a dependent variable see Table~\ref{app_tab:f1_ols_full}. $\text{*}\,p < 0.05$, Benjamini-Hochberg adjusted.}
    \label{tab:accuracy_ols}
\end{table}

\begin{table}[ht!]
    \centering
    \footnotesize
    \begin{tblr}{
      colsep=8pt,
      colspec={Q[1.5mm, colsep=0pt] Q[l, font=\bfseries, colsep=4pt] | X[l] X[l] X[l] }, 
      row{1}={font=\bfseries},
      rowsep=2pt,
      width=\linewidth,
      hline{2,3,5,8}={black},
      hborder{2,3,5,8}={belowspace=3pt}
    }
& & {ANES\\2016} & {GLES\\2017} & {GLES\\2025} \\
\SetCell{bg=tab_blue} & Intercept & .146* & .228* & .255* \\
\SetCell{bg=tab_blue2} & First-Token Restricted & .068* & .105* & .039 \\
\SetCell{bg=tab_blue3} & Answer Prefix & .032 & -.058 & -.087* \\
\SetCell{bg=tab_orange2} & Restricted Choice & .038 & -.082* & -.121* \\
\SetCell{bg=tab_orange2} & Restricted Reasoning & .050* & -.112* & -.148* \\
\SetCell{bg=tab_orange3} & Verbalized Distribution & \textbf{-.040*} & \textbf{-.131*} & \underline{-.150*} \\
\SetCell{bg=tab_green} & Open-Ended Classif. & .074* & -.093* & -.140* \\
\SetCell{bg=tab_green2} & Open-Ended Distrib. & .024 & \underline{-.129*} & \textbf{-.155*} \\
    \end{tblr}
    \caption{\textbf{Regression Coefficients for Subpopulation-Level Jensen-Shannon Divergence $(\downarrow)$.}
    OLS regression for each dataset, with Jensen-Shannon divergence $(\downarrow)$ as the dependent variable for the ANES and GLES datasets. Results with more than 10\% invalid values were excluded.  We use Survey Response Generation Method, response option variant, and LLM as independent variables. We show coefficients for Survey Response Generation Methods (Reference: First-Token Probabilities~{\color{tab_blue}$\blacksquare$}). For total variation distance as a dependent variable, see Table~\ref{app_tab:tv_ols_full}. $\text{*}\,p < 0.05$, Benjamini-Hochberg adjusted.}
    \label{tab:jsd_ols}
\end{table}

\begin{table}[t!]
    \centering
    \footnotesize
    \begin{tblr}{
      colsep=3pt,
      colspec={Q[1.5mm, colsep=0pt] Q[l, font=\bfseries, colsep=4pt] | X[l] X[l] X[l] X[l]}, 
      row{1}={font=\bfseries},
      rowsep=2pt,
      width=\linewidth,
      hline{2,3,5,8}={black},
      hborder{2,3,5,8}={belowspace=3pt}
    }
& & {ANES\\2016} & {GLES\\2017} & {GLES\\2025} & {ATP\\2021}\\
\SetCell{bg=tab_blue} & Intercept & .759* & .779* & .773* & .651* \\
\SetCell{bg=tab_blue2} & First-Token Restricted & \textbf{.124*} & -.058* & .018 & -.028 \\
\SetCell{bg=tab_blue3} & Answer Prefix & -.023 & -.101* & -.055 & -.030 \\
\SetCell{bg=tab_orange} & Restricted Choice & .111* & \underline{.053*} & .105* & .001 \\
\SetCell{bg=tab_orange2} & Restricted Reasoning & .117* & \textbf{.065*} & \underline{.117*} & \textbf{.059*} \\
\SetCell{bg=tab_orange3} & Verbalized Distribution & .098* & .046* & .105* & .013 \\
\SetCell{bg=tab_green} & Open-Ended Classif. & \underline{.123*} & .052* & \textbf{.123*} & \underline{.047*} \\
\SetCell{bg=tab_green2} & Open-Ended Distrib. & .103* & .040* & .109* & .037* \\
    \end{tblr}
    \caption{\textbf{Regression Coefficients for Subpopulation-Level Distance Correlation $(\uparrow)$.}
    OLS regression for each dataset, with Distance Correlation $(\uparrow)$ as the dependent variable. Results with more than 10\% invalid values were excluded. We use Survey Response Generation Method, response option variant, and LLM as independent variables. We show coefficients for Survey Response Generation Methods (Reference: First-Token Probabilities~{\color{tab_blue}$\blacksquare$}). For total variation distance as a dependent variable, see Table~\ref{app_tab:tv_ols_full}. $\text{*}\,p < 0.05$, Benjamini-Hochberg adjusted.}
    \label{tab:dcor_ols}
\end{table}

\begin{table*}[t!]
    \centering
    \small
    \begin{tblr}{
      colsep=8pt,
      colspec={Q Q | Q Q Q Q},
      rowsep=1.5pt,
      column{1,2}={font=\bfseries},
      row{1}={font=\bfseries},
    }
 &  & ANES 2016 & GLES 2017 & GLES 2025 & ATP 2021 \\
\hline \hborder{belowspace=3pt}
Intercept & & .343** & .037* & .050 & .107** \\
\hline \hborder{belowspace=3pt}
\SetCell[r=7]{h} {Survey Response\\Generation Method} & First-Token Restricted & .082** & .051* & .066 & -.032** \\
 & Answer Prefix & .013 & .069** & .059 & -.021* \\
 & Restricted Choice & .148** & .242** & .218** & .015 \\
 & Restricted Reasoning & .138** & .219** & .196** & .052** \\
 & Verbalized Distribution & .118** & .233** & .216** & .011 \\
 & Open-Ended Classif. & .127** & .215** & .184* & .037** \\
 & Open-Ended Distrib. & .114** & .212** & .177* & .018* \\
\hline \hborder{belowspace=3pt}
\SetCell[r=3]{h} {Response Option\\Variant} & Full Text, Reversed & -.004 & .011 & .002 & -.000 \\
 & Indexed & -.042** & -.020 & -.009 & .006 \\
 & Indexed, Reversed & -.054** & -.016 & -.016 & .005 \\
\hline \hborder{belowspace=3pt}
\SetCell[r=9]{h} {Model} & Llama 3B & -.020 & .007 & -.008 & .005 \\
 & Llama 70B & -.003 & .218** & .214** & .023* \\
 & OLMo 1B & -.073** & -.080** & -.086** & -.040** \\
 & OLMo 7B & .027 & .008 & -.006 & .022* \\
 & OLMo 32B & .061** & .155** & .179** & .034** \\
 & Qwen 8B & -.018 & .105** & .128** & .014 \\
 & Qwen 8B (R) & -.018 & .075** & .107** & .042** \\
 & Qwen 32B & .083** & .218** & .227** & .055** \\
 & Qwen 32B (R) & .038 & .160** & .196** & .091** \\
    \end{tblr}
    \caption{\textbf{Regression Coefficients for Individual-Level Alignment $(\uparrow)$.} OLS regression per dataset with macro avg.\ F1-score as the dependent variable (higher is better). We use Survey Response Generation Method, response option variant, and LLM name as independent variables. Reference: First-Token Probabilities Method + Full Text response options + Llama 8B. We do not include interactions to mitigate multicollinearity, and use clustered standard errors across seeds $\times$ decoding strategies to appropriately reflect the repeated-measures structure of our evaluation. $\text{*}\,p<0.05,\ \text{**}\,p < 0.01$, Benjamini-Hochberg adjusted.}
    \label{app_tab:f1_ols_full}
\end{table*}

\begin{table*}[t!]
    \centering
    \small
    \begin{tblr}{
      colsep=8pt,
      colspec={Q Q | X X X | X},
      rowsep=1.5pt,
      width=\textwidth,
      column{1,2}={font=\bfseries},
      row{1}={font=\bfseries},
    }
 &  & ANES 2016 & GLES 2017 & GLES 2025 & ATP 2021 \\
\hline \hborder{belowspace=3pt}
Intercept & & .322** & .476** & .578** & .070** \\
\hline \hborder{belowspace=3pt}
\SetCell[r=7]{h} {Survey Response\\Generation Method} & First-Token Restricted & .105** & .044 & -.069 & .008 \\
 & Answer Prefix & .049 & -.120** & -.202** & .002 \\
 & Restricted Choice & .045 & -.165** & -.287** & .012* \\
 & Restricted Reasoning & .061* & -.197** & -.312** & -.010* \\
 & Verbalized Distribution & -.028 & -.219** & -.296** & -.016** \\
 & Open-Ended Classif. & .072** & -.174** & -.306** & -.001 \\
 & Open-Ended Distrib. & .038 & -.216** & -.319** & -.006 \\
\hline \hborder{belowspace=3pt}
\SetCell[r=3]{h} {Response Option\\Variants} & Full Text, Reversed & .003 & -.006 & .038* & .001 \\
 & Indexed & .011 & .003 & -.001 & .002 \\
 & Indexed, Reversed & .037** & .012 & .026 & .002 \\
\hline \hborder{belowspace=3pt}
\SetCell[r=9]{h} {Model} & Llama 3B & -.049* & .034 & .069** & .002 \\
 & Llama 70B & -.047* & -.089** & -.133** & .017** \\
 & OLMo 1B & -.022 & .108** & .106** & .026** \\
 & OLMo 7B & -.060** & .064** & .069** & .009* \\
 & OLMo 32B & -.064** & -.073** & -.116** & .014** \\
 & Qwen 8B & .032 & .020 & -.056** & .022** \\
 & Qwen 8B (R) & -.010 & -.013 & -.024 & .013* \\
 & Qwen 32B & -.070** & -.112** & -.174** & .006* \\
 & Qwen 32B (R) & -.049** & -.069** & -.094** & -.005 \\
    \end{tblr}
    \caption{\textbf{Regression Coefficients for Subpopulation-Level Alignment $(\downarrow)$.} OLS regression per dataset with total variation distance (lower is better) as the dependent variable for the ANES and GLES datasets and 1-Wasserstein distance (lower is better) as the dependent variable for the ATP 2021 dataset. We use Survey Response Generation Method, response option variant, and LLM name as independent variables. Reference: First-Token Probabilities Method + Full Text response options + Llama 8B. We do not include interactions to mitigate multicollinearity, and use clustered standard errors across seeds $\times$ decoding strategies to appropriately reflect the repeated-measures structure of our evaluation. $\text{*}\,p<0.05,\ \text{**}\,p < 0.01$, Benjamini-Hochberg adjusted.}
    \label{app_tab:tv_ols_full}
\end{table*}

\begin{table*}[t!]
    \centering
    \small
    \begin{tblr}{
      colsep=5pt,
      colspec={Q Q | X[c] X[c] | X[c] X[c]},
      rowsep=1.5pt,
      width=\textwidth,
      column{1,2}={font=\bfseries},
      row{1}={c, font=\bfseries},
    }
& & \SetCell[c=2]{}{Individual-Level} & & \SetCell[c=2]{}{Subpopulation-Level} & \\
& & {\textbf{Accuracy}\\(Macro Avg.\\F1-Score)} & {\textbf{Robustness}\\(Fleiss' $\kappa$\\Across Scales)} & {\textbf{Alignment}\\(Total Var./\\1-Wasserst.)} & {\textbf{Global Align.}\\(Distance\\Correlation)} \\
\hline \hborder{belowspace=3pt}
Intercept & & -1.647** & -1.183** & -0.930** & -1.150** \\
\hline \hborder{belowspace=3pt}
\SetCell[r=7]{h} {Survey Response\\Generation Method} & First-Token Restricted & 0.751** & 0.461* & -0.029 & 0.613* \\
 & Answer Prefix & 0.415 & 0.009 & 0.667* & -0.073 \\
 & Restricted Choice & 1.535** & 1.178** & 0.845** & 1.138** \\
 & Restricted Reasoning & 1.576** & 1.097** & 1.090** & 1.302** \\
 & Verbalized Distribution & 1.443** & 0.655** & 1.433** & 1.104** \\
 & Open-Ended Classif. & 1.440** & 1.382** & 0.882** & 1.281** \\
 & Open-Ended Distrib. & 1.273** & 0.930** & 1.097** & 1.104** \\
\hline \hborder{belowspace=3pt}
\SetCell[r=3]{h} {Dataset} & ATP 2021 & 0.035 & 0.039 & 0.010 & 0.037 \\
 & GLES 2017 & 0.002 & -0.000 & -0.002 & 0.004 \\
 & GLES 2025 & 0.036 & 0.038 & 0.030 & 0.029 \\
\hline \hborder{belowspace=3pt}
\SetCell[r=3]{h} {Response Option\\Variants} & Full Text, Reversed & -0.034 & -0.046 & -0.083* & 0.049 \\
 & Indexed & -0.148** & -0.062* & -0.112** & -0.098** \\
 & Indexed, Reversed & -0.175** & -0.058* & -0.263** & -0.085* \\
\hline \hborder{belowspace=3pt}
\SetCell[r=9]{h} {Model} & Llama 3B & -0.145* & -0.477** & -0.113 & -0.404** \\
 & Llama 70B & 0.850** & 0.971** & 0.441** & 0.341** \\
 & OLMo 1B & -0.746** & -1.306** & -0.541** & -0.957** \\
 & OLMo 7B & 0.175** & -0.388** & -0.063 & -0.010 \\
 & OLMo 32B & 0.850** & 0.851** & 0.446** & 0.585** \\
 & Qwen 8B & 0.476** & 0.560** & -0.170** & 0.278** \\
 & Qwen 8B (R) & 0.663** & 0.817** & 0.171** & 0.437** \\
 & Qwen 32B & 1.103** & 0.916** & 0.703** & 0.594** \\
 & Qwen 32B (R) & 1.163** & 1.069** & 0.623** & 0.679** \\
    \end{tblr}
    \caption{\textbf{Regression Coefficients For Evaluation Metrics (normalized, $\uparrow$).} OLS regression on evaluation outcomes for individual-level accuracy, and robustness (Fleiss' $\kappa$), as well as subpopulation-level alignment ($1 -$ total variation distance / 1-Wasserstein distance) and global alignment (distance correlation). Results from each metric are z-score normalized separately on each dataset. We use dataset, Survey Response Generation Method, response option variant, and LLM as independent variables, and do not include interactions. Reference: ANES 2016 + First-Token Probabilities~{\color{tab_blue}$\blacksquare$} + Full Text response options + Llama 8B. Results with more than 10\% invalid values were excluded. \textbf{Restricted Generation Methods~{\color{tab_orange}$\blacksquare$}{\color{tab_orange2}$\blacksquare$}{\color{tab_orange3}$\blacksquare$} consistently lead to significant improvements} $\text{*}\,p<0.05,\ \text{**}\,p < 0.01$, Benjamini-Hochberg adjusted.}
    \label{app_tab:aggregate_ols_full}
\end{table*}

\begin{figure*}[t!]
    \centering
    \hfill
    \begin{subfigure}[][0.85\textheight][t]{0.445\textwidth}
        \centering
        \includegraphics[width=\linewidth]{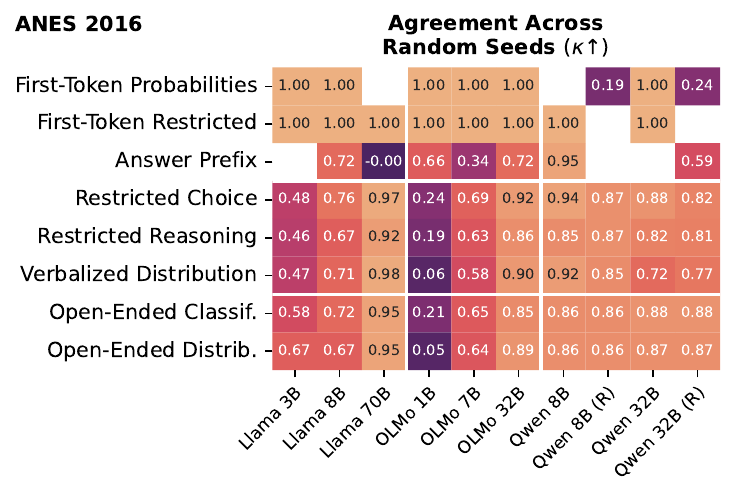}
        \includegraphics[width=\linewidth]{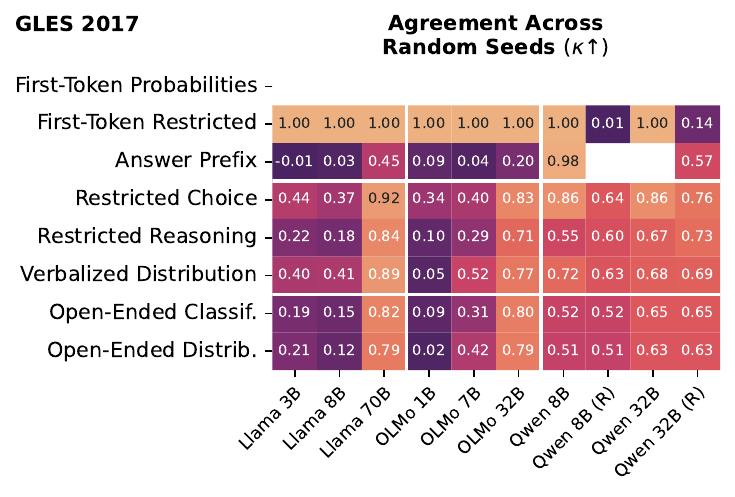}
        \includegraphics[width=\linewidth]{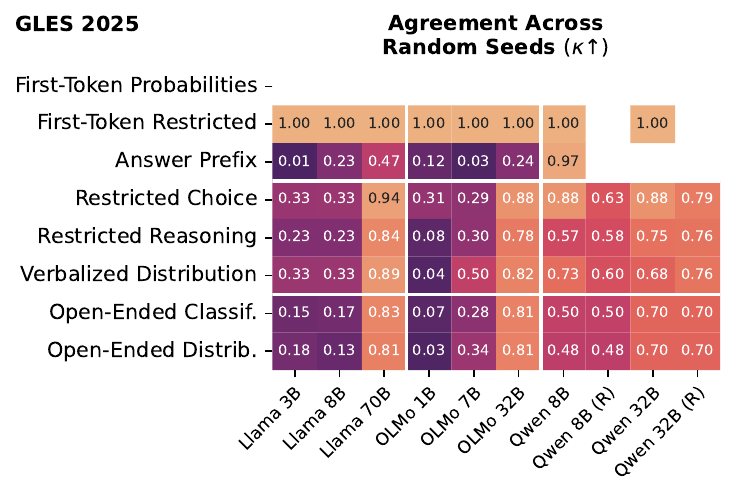}
        \hfill
        \caption{\textbf{Individual-Level Agreement Across Seeds.} The First-Token Probability Method and the First-Token Restricted Method achieve perfect agreement across all non-reasoning models, as the output probabilities of the first token are deterministic given the same prompt. No agreement across seeds is calculated for the ATP 2021 dataset, as we evaluated this dataset with only 1 seed to save computational resources.}
        \label{app_fig:robustness_seeds}
    \end{subfigure}
    \hfill
    \begin{subfigure}[][0.85\textheight][t]{0.445\textwidth}
        \centering
        \includegraphics[width=\linewidth]{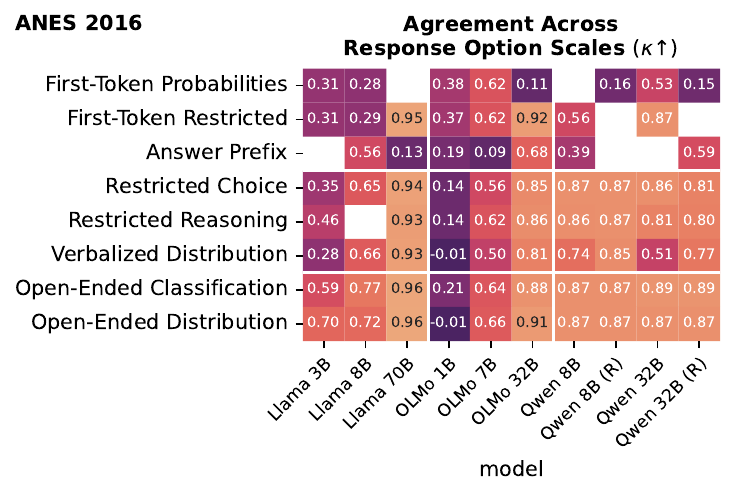}
        \includegraphics[width=\linewidth]{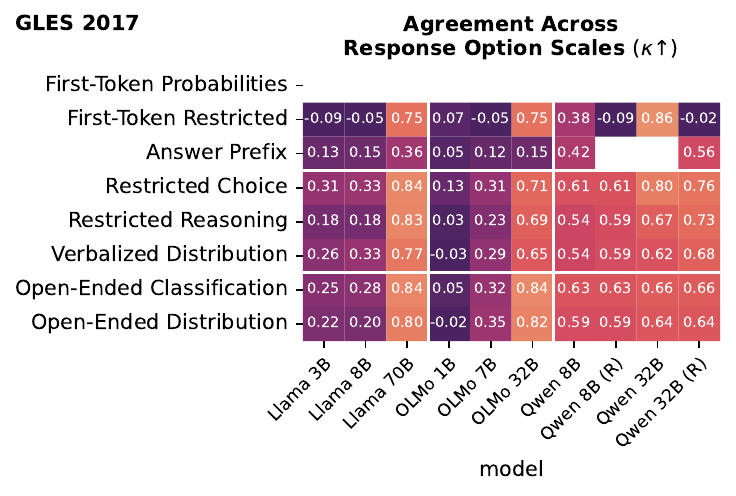}
        \includegraphics[width=\linewidth]{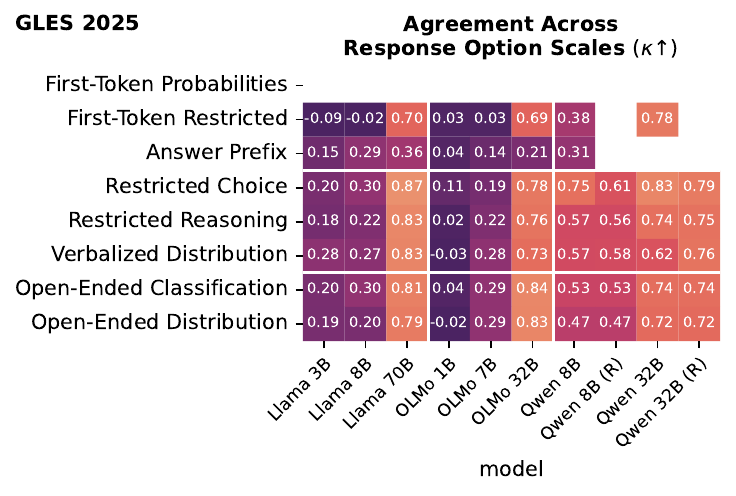}
        \includegraphics[width=\linewidth]{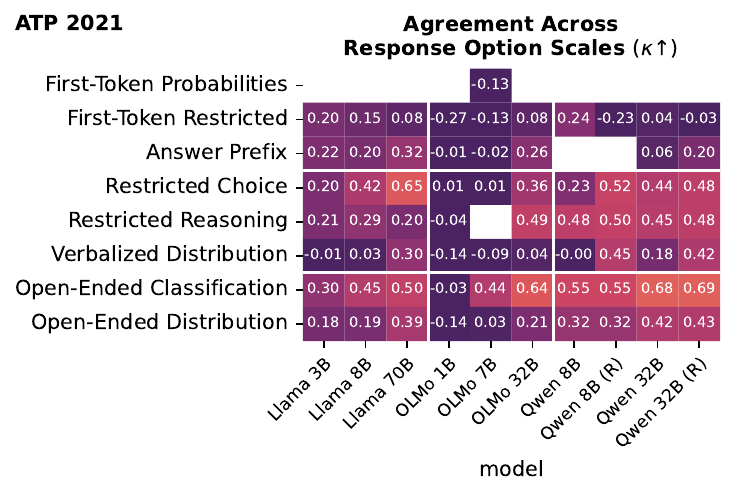}
        \caption{\textbf{Individual-Level Agreement Across Response Option Scales.} A sufficient agreement is often desirable, as perturbations in the response options scales should not impact the response that is generated by a model.}
        \label{app_fig:robustness_scales}
    \end{subfigure}
    \hfill
    \caption{\textbf{Individual-Level Agreement.} Mean Fleiss's $\kappa\,(\uparrow)$, results with more than 10\% invalid values excluded. Perfect agreement across seeds or scales might not be considered desirable, as it indicates overly confident individual-level predictions given the variance in human survey responses. For an evaluation of individual-level calibration, see also Figure~\ref{app_fig:calibration}. }
    \label{app_fig:robustness_seeds_scales}
\end{figure*}

\begin{figure*}[t!]
    \centering
    \begin{subfigure}[][0.4\textheight][t]{0.48\textwidth}
        \includegraphics[width=\linewidth]{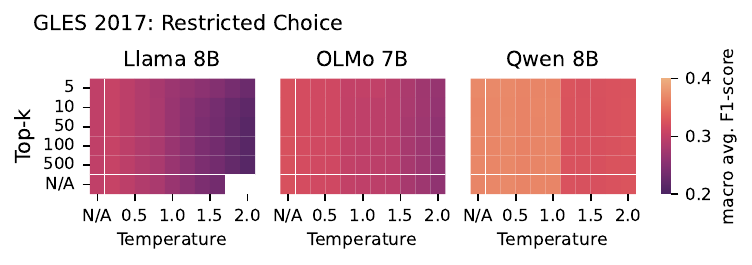}
        \includegraphics[width=\linewidth]{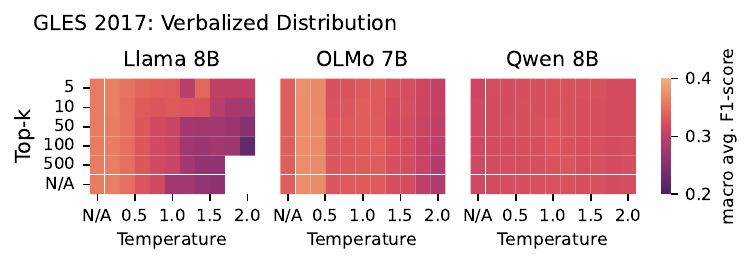}
        \includegraphics[width=\linewidth]{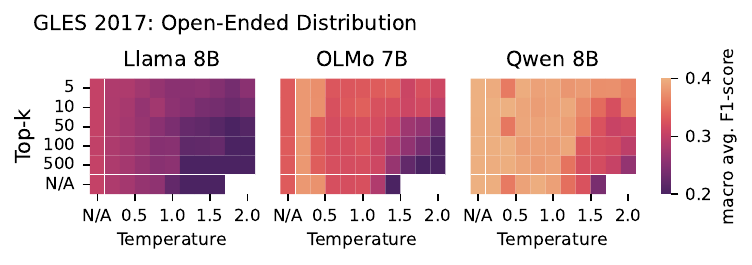}
        \caption{\textbf{Decoding Strategy Impacts Individual-Level Alignment.} Macro avg.\ F1-score $(\uparrow)$---results generally degrade with increasing temperature.}
        \label{app_fig:temp_topk_accuracy}
    \end{subfigure}
    \hfill
    \vspace{0.2cm}
    \begin{subfigure}[][0.4\textheight][t]{0.48\textwidth}
        \includegraphics[width=\linewidth]{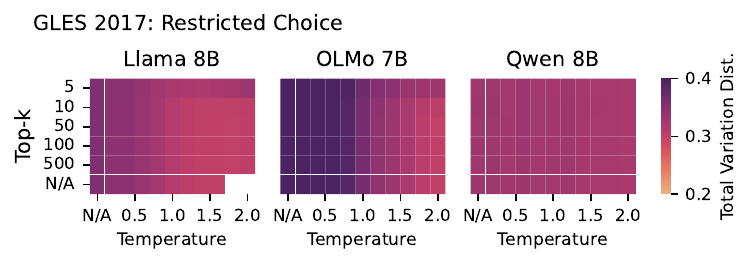}
        \includegraphics[width=\linewidth]{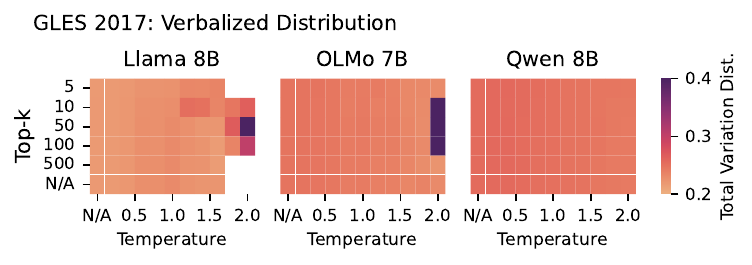}
        \includegraphics[width=\linewidth]{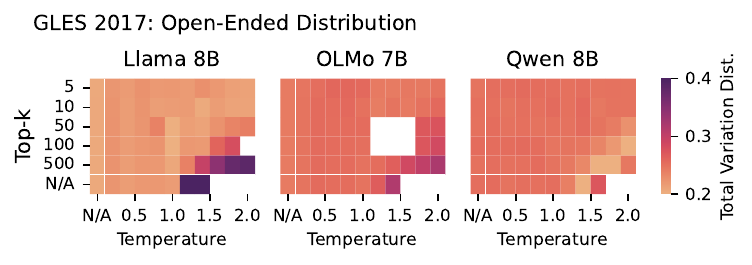}
        \caption{\textbf{Decoding Strategy Impacts Subpopulation-Level Alignment.} Total variation distance $(\downarrow)$---results improve for \texttt{Llama 8B} and \texttt{OLMo 7B} with the Restricted Choice Method, but are otherwise mostly stable.}
        \label{app_fig:temp_topk_alignment}
    \end{subfigure}
    \caption{\textbf{Decoding Strategies.} For 3 Survey Response Generation Methods (Restricted Choice, Verbalized Distribution, and Open-Ended Distribution) and 3 medium-size LLMs (\texttt{Llama 8B}, \texttt{OLMo 7B}, and \texttt{Qwen 8B}), we investigate a diverse range of \texttt{temperature} and \texttt{top-k} values during decoding. \texttt{N/A} stands for greedy decoding or full vocabulary respectively. Results with more than 10\% invalid values excluded.}
    \label{app_fig:temp_topk}
\end{figure*}

\begin{figure*}[t!]
    \centering
    \includegraphics[width=0.45\linewidth]{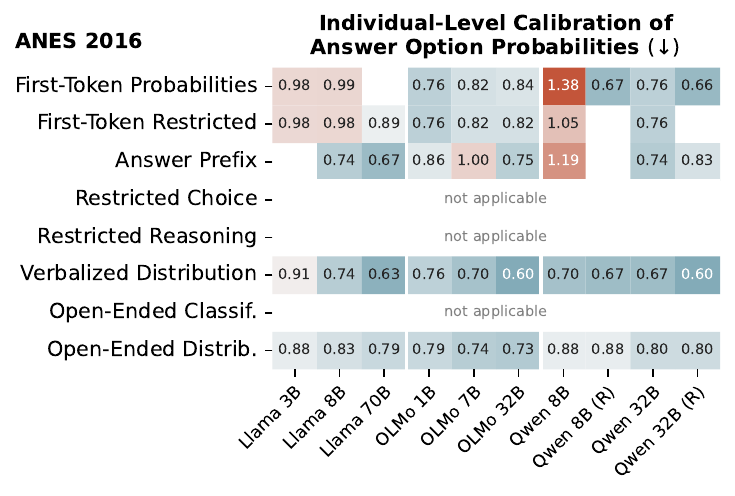}
    \includegraphics[width=0.45\linewidth]{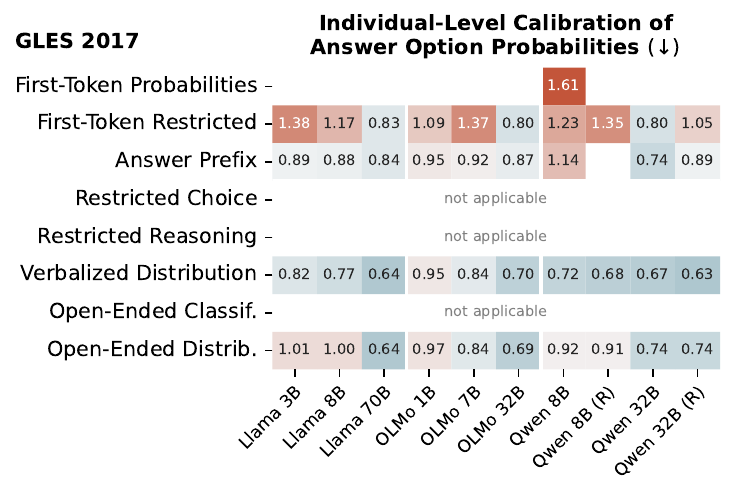}
    \includegraphics[width=0.45\linewidth]{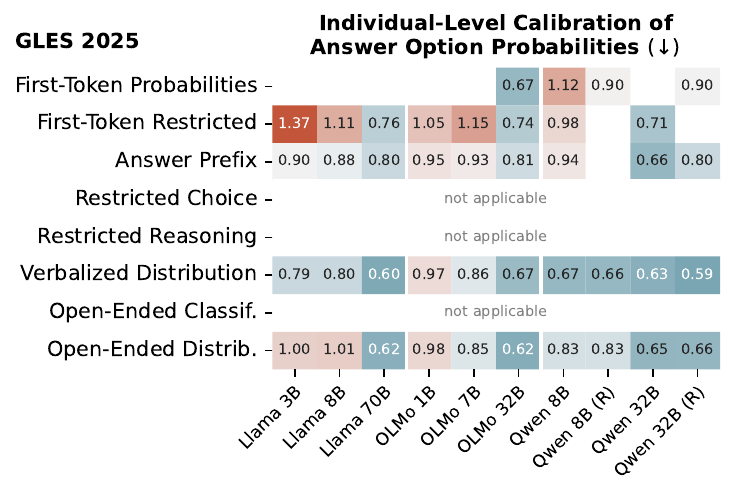}
    \includegraphics[width=0.45\linewidth]{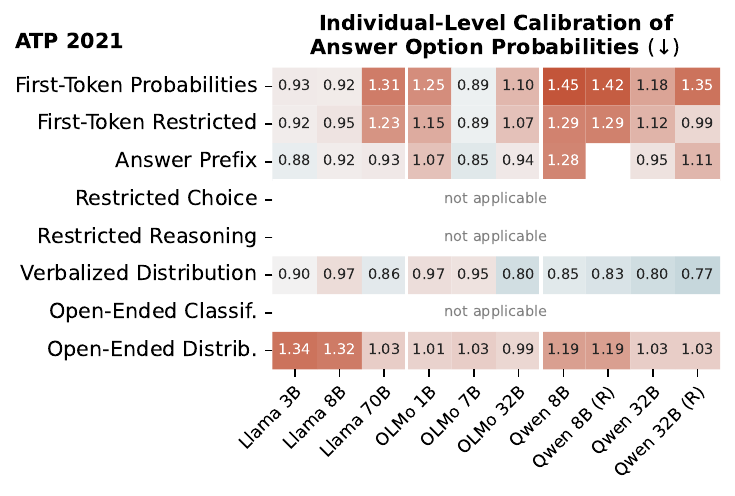}
    \caption{\textbf{Individual-Level Calibration $(\downarrow)$.} Mean Brier score $(\downarrow)$ across all response options, results with more than 10\% invalid values excluded. For Survey Response Generation Methods that generate individual-level distributions across response options, we can evaluate whether high ``confidence'' of a model in a response option corresponds with correct predictions on the individual level. The Brier score calculates individual-level calibration as the mean squared error between the confidences and the one-hot encoded human survey responses. Well-calibrated Survey Response Generation Methods are desirable, as they accurately capture individual-level prediction uncertainty. We find that larger models are generally better calibrated than smaller ones. Token Probability-Based Methods can be poorly calibrated for most models, while the \textbf{Verbalized Distribution Method leads to the best individual-level calibration.}}
    \label{app_fig:calibration}
\end{figure*}


\begin{figure*}
    \centering
    \includegraphics[width=0.95\linewidth]{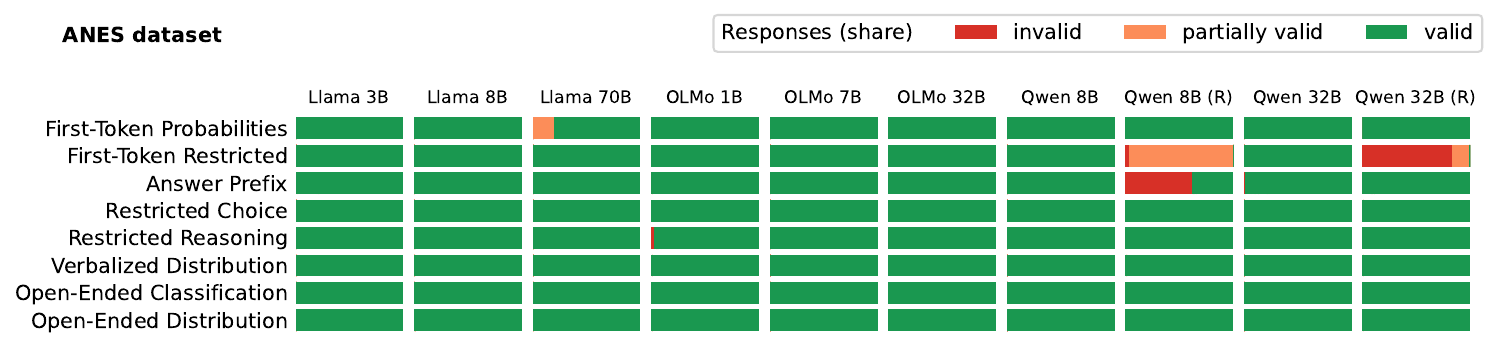}
    \includegraphics[width=0.95\linewidth]{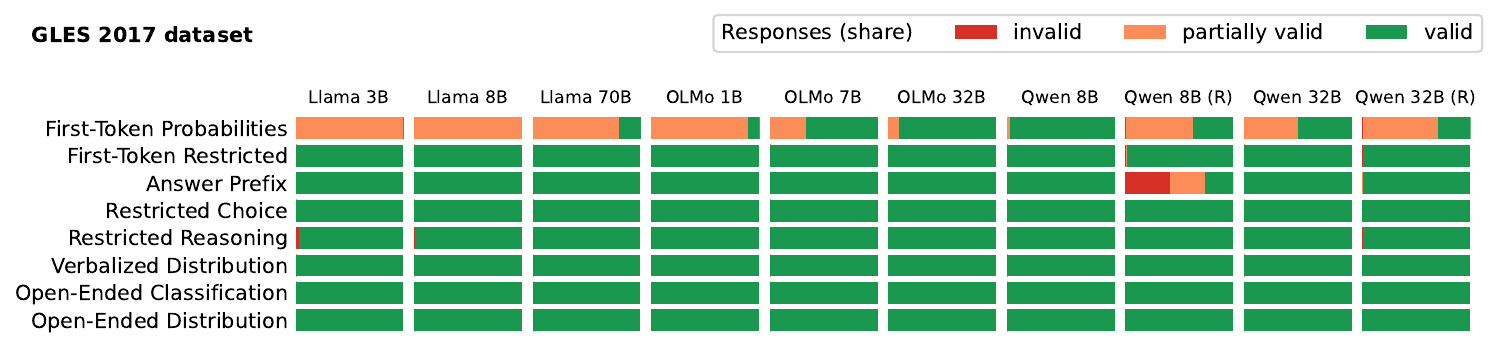}
    \includegraphics[width=0.95\linewidth]{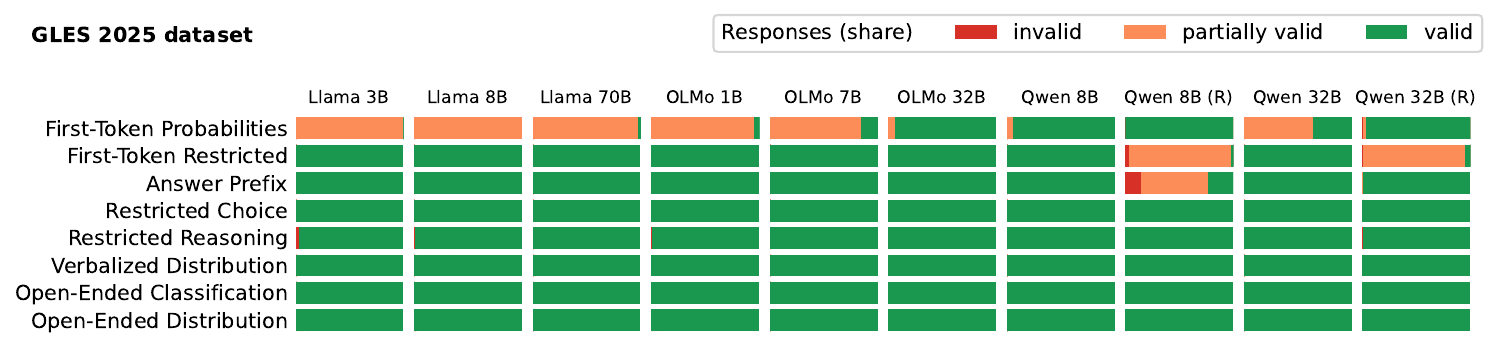}
    \includegraphics[width=0.95\linewidth]{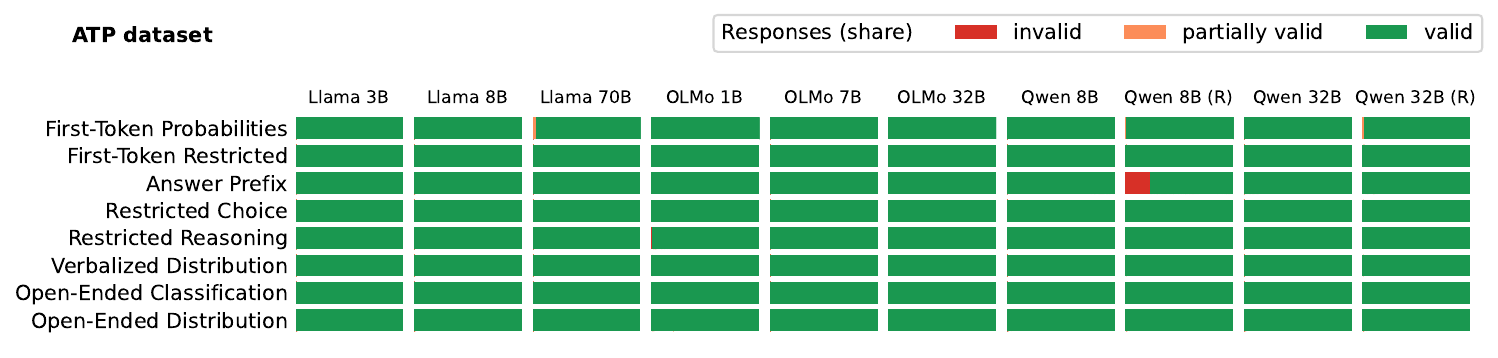}
    \caption{\textbf{Fractions of LLM Responses That Are Invalid.} We count model output that does not contain any of the possible closed-ended response options as \textit{invalid.} For Survey Response Generation Methods that generate distributions over possible response options (see Table~\ref{tab:methods_overview}), we count model output that only contains probabilities for some of the response options as \textit{partially valid.}}
    \label{app_fig:invalid_responses}
\end{figure*}

\begin{table*}[hb!]
    \centering
    \scriptsize
    \begin{tblr}{
        width=\textwidth,
        colspec={Q | X},
        column{2}={2-Z}{font=\ttfamily},
        row{1}={font=\bfseries},
        column{1}={font=\bfseries},
        hline{2-Y}={black},
        hborder{2-Y}={belowspace=5pt},
    }
{Survey Response\\Generation Method} & {Formatting Instructions\\(System Prompt)}\\

{Token Probability\\-Based Methods} &
You are a political scientist predicting responses to the following question: \{QUESTION\}.
These are the possible answer options: \{RESPONSE OPTIONS\}.
You only respond with the most probable answer option. \\

Restricted Choice &
{You are a political scientist predicting responses to the following question: \{QUESTION\}.
These are the possible answer options: \{RESPONSE OPTIONS\}.
You only respond with the most probable answer option in the following JSON format: \\
\textasciigrave\textasciigrave\textasciigrave json \\
\{ \\
  "answer": \{RESPONSE OPTIONS\} \\
\} \\
\textasciigrave\textasciigrave\textasciigrave }\\

Restricted Reasoning &
{You are a political scientist predicting responses to the following question: \{QUESTION\}.
These are the possible answer options: \{RESPONSE OPTIONS\}.
You always reason about the possible answer options first. \\
You respond with your reasoning and the most probable answer option in the following JSON format: \\
\textasciigrave\textasciigrave\textasciigrave json \\
\{ \\
  "reasoning": <your reasoning about the answer options>, \\
  "answer": <\{RESPONSE OPTIONS\}> \\
\} \\
\textasciigrave\textasciigrave\textasciigrave }\\

Verbalized Distribution &
{You are a political scientist predicting responses to the following question: \{QUESTION\}.
These are the possible answer options: \{RESPONSE OPTIONS\}.
You only respond with a probability for each answer option in the following JSON format: \\
\textasciigrave\textasciigrave\textasciigrave json \\
\{ \\
  \{RESPONSE OPTION 1\}: <probability>, \\
  \{RESPONSE OPTION 2\}: <probability>, \\
  \{...\} \\
\} \\
\textasciigrave\textasciigrave\textasciigrave }\\

{Open Generation Methods\\(Step 1: Open Generation)} &
You are a political scientist predicting responses to the following question: \{QUESTION\}. \\

{Open-Ended Classification\\(Step 2: Classification\textsuperscript{1})}&
{You are an expert annotator.
These are the possible labels: \{RESPONSE OPTIONS\}.
You only respond with the most probable label in the following JSON format: \\
\textasciigrave\textasciigrave\textasciigrave json \\
\{ \\
  "answer": <\{RESPONSE OPTIONS\}> \\
\} \\
\textasciigrave\textasciigrave\textasciigrave }\\

{Open-Ended Distribution\\(Step 2: Classification\textsuperscript{1})} &
{You are an expert annotator.
These are the possible labels: \{RESPONSE OPTIONS\}.
You only respond with a probability for each answer option in the following JSON format: \\
\textasciigrave\textasciigrave\textasciigrave json \\
\{ \\
  \{RESPONSE OPTION 1\}: <probability>, \\
  \{RESPONSE OPTION 2\}: <probability>, \\
  \{...\} \\
\} \\
\textasciigrave\textasciigrave\textasciigrave }\\

    \end{tblr}
    \caption{\textbf{Formatting Instructions (System Prompts) for English---ANES 2016 \& ATP 2021 Datasets.} The \texttt{QUESTION} and \texttt{RESPONSE OPTIONS} are identical to what was used by~\citet{argyle_out_2023} and \citet{santurkar_whose_2023} respectively. \textsuperscript{1}Additional user prompts for the classification step are shown in Table~\ref{tab:EN_class_prompts}.}
    \label{tab:EN_system_prompts}
\end{table*}

\begin{table*}[hb!]
    \centering
    \scriptsize
    \begin{tblr}{
        width=\textwidth,
        colspec={Q | X},
        column{2}={2-Z}{font=\ttfamily},
        row{1}={font=\bfseries},
        column{1}={font=\bfseries},
        hline{2-Y}={black},
        hborder{2-Y}={belowspace=5pt},
    }
{Survey Response\\Generation Method} & {Classification Instructions\\(User Prompt)}\\

{Open-Ended Classification /\\Open-Ended Distribution\\(Step 2: Classification)} &
{Below is a STATEMENT that responds to the question: \{QUESTION\}
Does the following STATEMENT express preference for exactly one of the following answer options: <\{RESPONSE OPTIONS\}>? If yes, which option is preferred?

STATEMENT: \{OPEN-ENDED MODEL RESPONSE\}}\\
    \end{tblr}
    \caption{\textbf{Classification Instructions (User Prompts) for English---ANES 2016 \& ATP 2021 Datasets.} These prompts are only used in the second, classification step of the Open-Ended Classification and the Open-Ended Distribution Methods. The user prompts for all other Survey Response Generation Methods, as well as for the first, open-ended step of the Open Generation Methods are shown in Table~\ref{tab:persona_prompts}.}
    \label{tab:EN_class_prompts}
\end{table*}

\begin{table*}[hb!]
    \centering
    \scriptsize
    \begin{tblr}{
        width=\textwidth,
        colspec={Q | X},
        column{2}={2-Z}{font=\ttfamily},
        row{1}={font=\bfseries},
        column{1}={font=\bfseries},
        hline{2-Y}={black},
        hborder{2-Y}={belowspace=5pt},
    }
{Survey Response\\Generation Method} & {Formatting Instructions\\(System Prompt)}\\

{Token Probability\\-Based Methods} &
Du bist ein Politikwissenschaftler, der Antworten auf die folgende Frage vorhersagt: \{QUESTION\}.
Dies sind die möglichen Antwortoptionen: \{RESPONSE OPTIONS\}.
You only respond with the most probable answer option. \\

Restricted Choice &
{Du bist ein Politikwissenschaftler, der Antworten auf die folgende Frage vorhersagt: \{QUESTION\}.
Dies sind die möglichen Antwortoptionen: \{RESPONSE OPTIONS\}.
Du antwortest ausschließlich mit der wahrscheinlichsten Antwortoption im folgenden JSON-Format: \\
\textasciigrave\textasciigrave\textasciigrave json \\
\{ \\
  "antwort": \{RESPONSE OPTIONS\} \\
\} \\
\textasciigrave\textasciigrave\textasciigrave }\\

Restricted Reasoning &
{Du bist ein Politikwissenschaftler, der Antworten auf die folgende Frage vorhersagt: \{QUESTION\}.
Dies sind die möglichen Antwortoptionen: \{RESPONSE OPTIONS\}.
Du argumentierst immer zuerst über die möglichen Antwort-Optionen. \\
Du antwortest mit deiner Argumentation und der wahrscheinlichsten Antwort-Option im folgenden JSON-Format: \\
\textasciigrave\textasciigrave\textasciigrave json \\
\{ \\
  "argumentation": <deine Argumentation über die Antwort-Optionen>, \\
  "antwort": <\{RESPONSE OPTIONS\}> \\
\} \\
\textasciigrave\textasciigrave\textasciigrave }\\

Verbalized Distribution &
{Du bist ein Politikwissenschaftler, der Antworten auf die folgende Frage vorhersagt: \{QUESTION\}.
Dies sind die möglichen Antwortoptionen: \{RESPONSE OPTIONS\}.
Du antwortest ausschließlich mit einer Wahrscheinlichkeit für jede Antwort-Option im folgenden JSON-Format: \\
\textasciigrave\textasciigrave\textasciigrave json \\
\{ \\
  \{RESPONSE OPTION 1\}: <Wahrscheinlichkeit>, \\
  \{RESPONSE OPTION 2\}: <Wahrscheinlichkeit>, \\
  \{...\} \\
\} \\
\textasciigrave\textasciigrave\textasciigrave }\\

{Open Generation Methods\\(Step 1: Open Generation)} &
Du bist ein Politikwissenschaftler, der Antworten auf die folgende Frage vorhersagt: \{QUESTION\}. \\

{Open-Ended Classification\\(Step 2: Classification\textsuperscript{1})}&
{Du bist ein erfahrener Annotator.
Das sind die möglichen Labels: \{RESPONSE OPTIONS\}.
Du antwortest nur mit dem wahrscheinlichsten Label im folgenden JSON-Format: \\
\textasciigrave\textasciigrave\textasciigrave json \\
\{ \\
  "antwort": <\{RESPONSE OPTIONS\}> \\
\} \\
\textasciigrave\textasciigrave\textasciigrave }\\

{Open-Ended Distribution\\(Step 2: Classification\textsuperscript{1})} &
{Du bist ein erfahrener Annotator.
Das sind die möglichen Labels: \{RESPONSE OPTIONS\}.
Du antwortest nur mit einer Wahrscheinlichkeit für jede Antwortoption im folgenden JSON-Format: \\
\textasciigrave\textasciigrave\textasciigrave json \\
\{ \\
  \{RESPONSE OPTION 1\}: <Wahrscheinlichkeit>, \\
  \{RESPONSE OPTION 2\}: <Wahrscheinlichkeit>, \\
  \{...\} \\
\} \\
\textasciigrave\textasciigrave\textasciigrave }\\

    \end{tblr}
    \caption{\textbf{Formatting Instructions (System Prompts) for German---GLES 2017 \& GLES 2025 Datasets.} The \texttt{QUESTION} and \texttt{RESPONSE OPTIONS} are identical to what was used by~\citet{von_der_heyde_vox_2025}. \textsuperscript{1}Additional user prompts for the classification step are shown in Table~\ref{tab:DE_class_prompts}.}
    \label{tab:DE_system_prompts}
\end{table*}

\begin{table*}[hb!]
    \centering
    \scriptsize
    \begin{tblr}{
        width=\textwidth,
        colspec={Q | X},
        column{2}={2-Z}{font=\ttfamily},
        row{1}={font=\bfseries},
        column{1}={font=\bfseries},
        hline{2-Y}={black},
        hborder{2-Y}={belowspace=5pt},
    }
{Survey Response\\Generation Method} & {Classification Instructions\\(User Prompt)}\\

{Open-Ended Classification /\\Open-Ended Distribution\\(Step 2: Classification)} &
{Nachfolgend findest du eine AUSSAGE, die auf die Frage \{QUESTION\} antwortet.
Drückt die folgende AUSSAGE eine Präferenz für genau eine der folgenden Antwortoptionen aus: <\{RESPONSE OPTIONS\}>? Wenn ja, welche Option wird bevorzugt?

AUSSAGE: \{OPEN-ENDED MODEL RESPONSE\}}\\
    \end{tblr}
    \caption{\textbf{Classification Instructions (User Prompts) for German---GLES 2017 \& GLES 2025 Datasets.} These prompts are only used in the second, classification step of the Open-Ended Classification and the Open-Ended Distribution Methods. The user prompts for all other Survey Response Generation Methods, as well as for the first, open-ended step of the Open Generation Methods are shown in Table~\ref{tab:persona_prompts}.}
    \label{tab:DE_class_prompts}
\end{table*}

\begin{table*}[hb!]
    \centering
    \scriptsize
    \begin{tblr}{
        width=\textwidth,
        colspec={Q | X},
        column{2}={2-Z}{font=\ttfamily},
        row{1}={font=\bfseries},
        column{1}={font=\bfseries},
        hline{2-Y}={black},
        hborder{2-Y}={belowspace=5pt},
    }
{Dataset} & {Persona \& Question (User Prompt)}\\
{ANES 2016\\(English)} & {Racially, I am \{row.race\}.
\{"I like to discuss politics with my family and friends." if row.discuss\_politics == 'Yes' else "I never discuss politics with my family or friends."\} Ideologically, I am \{row.ideology\}. Politically, I am an \{row.party\}. I \{row.church\_goer\}. I am \{int(row.age)\} years old. I am a \{row.gender\}. I am \{row.political\_interest\} interested in politics. It makes me feel \{row.patriotism\} to see the American flag. I am from \{row.state\}.
\\ 
\\ In the 2016 presidential election, I voted for
} \\

{GLES 2017\\(German)} &  {Ich bin \{row.Alter\} Jahre alt und \{row.Geschlecht\}. Ich habe \{row.Bildung\}, ein \{row.Haushaltseinkommen\} monatliches Haushalts-Nettoeinkommen und bin \{row['Erwerbstätigkeit']\}. Ich bin \{row['Religiosität']\}. Politisch-ideologisch ordne ich mich \{row['Links-Rechts-Einstufung']\} ein. Ich identifiziere mich \{row['Parteiidentifikation Stärke']\} \{row['Parteiidentifikation']\}. Ich lebe in \{row['Ost-West']\}. Ich finde, die Regierung sollte die Einwanderung \{row['Zuwanderung']\} und \{row['Einkommensunterschiede verringern']\} ergreifen, um die Einkommensunterschiede zu verringern.
\\
\\ Habe ich bei der Bundestagswahl 2017 gewählt und wenn ja, welcher Partei habe ich meine Zweitstimme gegeben?} \\

{GLES 2025\\(German)} & --- same as for 2017 with only the year in the question changed --- \\

{ATP 2021\\(English)} &
{In politics today, I consider myself a \{row.POLPARTY\}. I would describe my political views as \{row.POLIDEOLOGY\}. My present religion is \{row.RELIG\}. I am \{row.RACE\}. The highest level of education I have completed is: \{row.EDUCATION\}. Last year, my total family income from all sources, before taxes was \{row.INCOME\}. I currently reside in the \{row.CREGION\}. I identify as \{row.SEX\}.
\\
\\ \{QUESTION\}
} \\

    \end{tblr}
    \caption{\textbf{Persona \& Question Prompts (User Prompts) for All Datasets.} All prompts closely follow the templates provided by~\citet{argyle_out_2023, von_der_heyde_vox_2025, santurkar_whose_2023}. During the evaluations, we omit sentences with missing data on at least one of the variables.}
    \label{tab:persona_prompts}
\end{table*}

\end{document}